\documentclass{elsart}

\usepackage{epsfig}

\newcommand{\Natural}{\mathop{\rm I\kern-.2emN}}
\newcommand{\Real}{\mathop{\rm I\kern-.2emR}}
\newcommand{\definition}[1]{\textbf{Definition:} #1\\}

\begin{document}

\begin{frontmatter}
\date{}

\title{Fitness Landscape of the Cellular Automata Majority Problem:
\\View from the "Olympus"}

\author[I3S]{S. Verel},
\author[I3S]{P. Collard},
\author[Tomassini]{M. Tomassini} and
\author[Vanneschi]{L. Vanneschi}

\address[I3S]{Laboratoire I3S, CNRS-University of Nice Sophia Antipolis, France}
\address[Tomassini]{Information Systems Department, University of Lausanne, Switzerland}
\address[Vanneschi]{Dipartimento di Informatica Sistemistica e Comunicazione, University of Milano-Bicocca, Italy}

\begin{abstract}
In this paper we study cellular automata (CAs) that perform the computational Majority task. 
This task is a good example of what the phenomenon of emergence in complex systems is. We take an interest in the
reasons that make this particular fitness landscape a difficult one. The
first goal is to study landscape as such, and thus it is ideally
independent from the actual heuristics used to search the space. However, a
second goal is to understand the features a good search technique for this
particular problem space should possess. We statistically quantify in
various ways the degree of difficulty of searching this landscape. Due to neutrality, investigations based on sampling techniques on the whole landscape are difficult to conduct. So, we go exploring the landscape
\textit{from the top}. Although it has been proved that no CA can perform the task
perfectly, several efficient CAs for this task have been found. Exploiting similarities between these CAs and symmetries in the landscape, we define the \textit{Olympus} landscape which is regarded as the "heavenly home" of the \textit{best local optima known} (blok). 
Then we measure several properties of this subspace. Although it is easier
to find relevant CAs in this subspace than in the overall landscape, there
are structural reasons that prevents a searcher from finding overfitted CAs
in the Olympus.
Finally, we study dynamics and performances of genetic algorithms on the \textit{Olympus} in order to
confirm our analysis and to find efficient CAs for the Majority problem with low computational cost.
\end{abstract}


\end{frontmatter}

\section{Introduction}
\label{intro}

Cellular automata (CAs) are discrete dynamical systems that have been studied
theoretically for years due to their architectural simplicity and the wide spectrum
of behaviors they are capable of \cite{chopard97,wolfram-book-02}.
CAs are capable of universal
computation and their time evolution can be complex.
But many CAs show simpler dynamical behaviors such as fixed points and cyclic 
attractors.
Here we study CAs that can be said to perform a simple ``computational'' task. One
such tasks is the so-called \textit{majority} or \textit{density} task in which a
two-state CA is to decide
whether the initial state contains more zeros than ones or \textit{vice versa}. In spite of
the apparent simplicity of the task, it is difficult for a local system as a CA as
it requires a coordination among the cells. As such, it is a perfect paradigm
of the phenomenon of \textit{emergence} in complex systems. That is, the task solution is
an emergent global property of a system of locally interacting agents.
Indeed, it has been proved that no CA can perform the task perfectly i.e., for 
any possible initial binary
configuration of states \cite{landbelew95}. However, several efficient CAs for the density task
have been found either by hand or by using heuristic methods, especially evolutionary
computation \cite{mitchelletal93,mitchelletal94a,sipper96epcm,andreetal96b,juille98,Breukelaar05}.
For a recent review of the work done on the problem in the last ten years
see \cite{crutch-mitch-das-03}.

All previous investigations have empirically shown that finding good CAs
for the majority task is very hard. In other words, the space of automata that are
feasible solutions to the task is a difficult one to search. However, there have been no
investigations, to our knowledge, of the reasons that make this particular fitness
landscape a difficult one. In this paper we try to statistically quantify in
various ways the degree of difficulty of searching the majority CA landscape. 
Our investigation is a study of the fitness landscape as such, 
and thus it is ideally independent from the actual heuristics used to search the space
provided that they use independent bit mutation as a search operator.
However, a second goal of this study is to understand the features a good
search technique for this particular problem space should possess.\\
The present study follows in the line of previous work by Hordijk \cite{hord-97}
for another interesting collective CA problem: the synchronization task \cite{dasetal95}.

The paper proceeds as follows. The next section summarizes
definitions and facts about CAs and the density task, including previous results obtained in 
building CAs for the task. A description of fitness landscapes and their statistical analysis follows.
This is followed by a detailed analysis of the majority problem fitness landscape. Next we identify
and analyze a particular subspace of the problem search space called the Olympus. Finally, we
present our conclusions and hints to further works and open questions.

\section{Cellular Automata and the Majority Problem}
\label{ca}

\subsection{Cellular Automata}

\begin{figure}[!ht]
\begin{center}
\begin{tabular}{ccc}
\mbox{
  \epsfig{figure=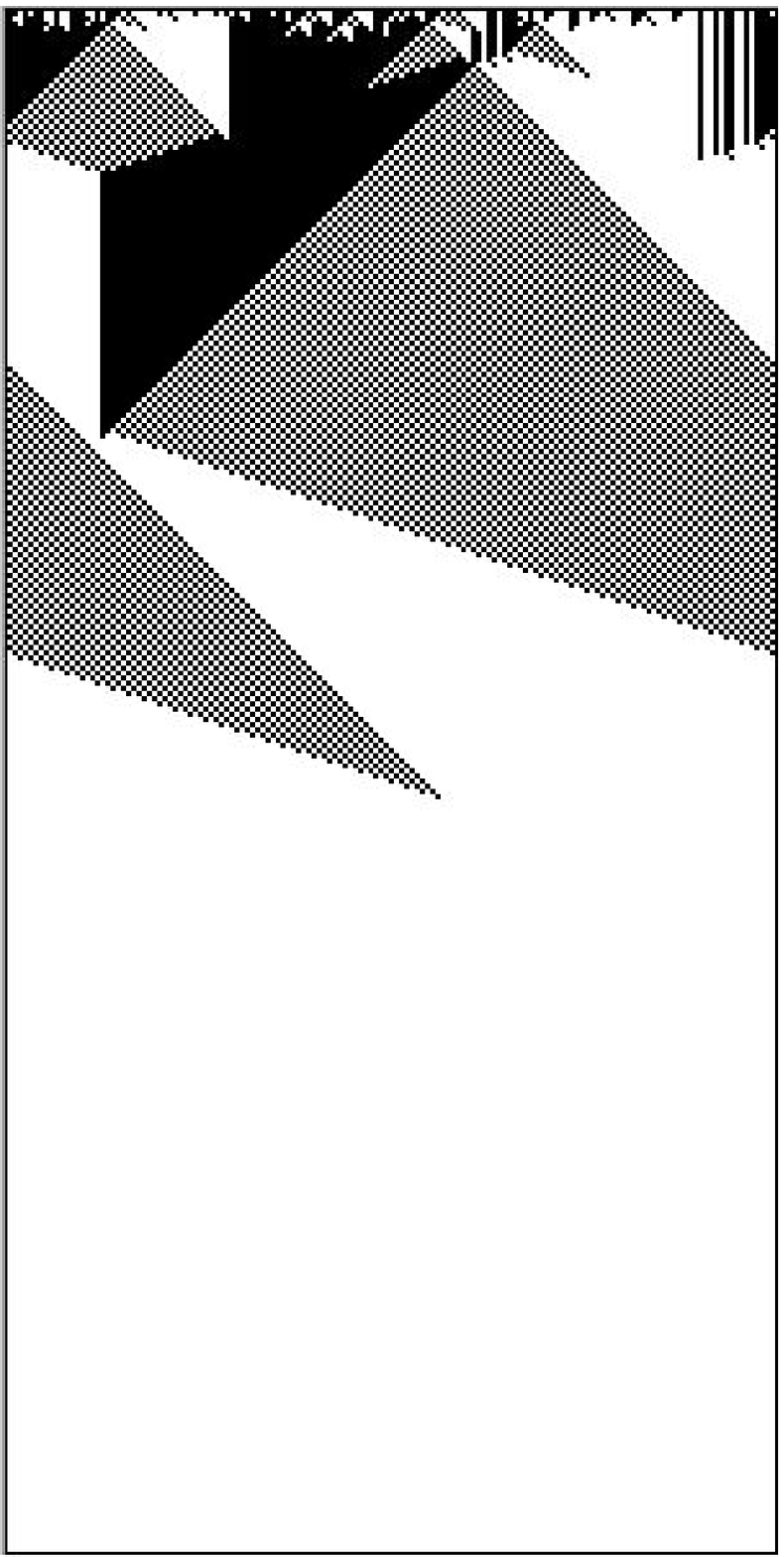,width=4.0cm,height=6cm} } & &
\mbox{
  \epsfig{figure=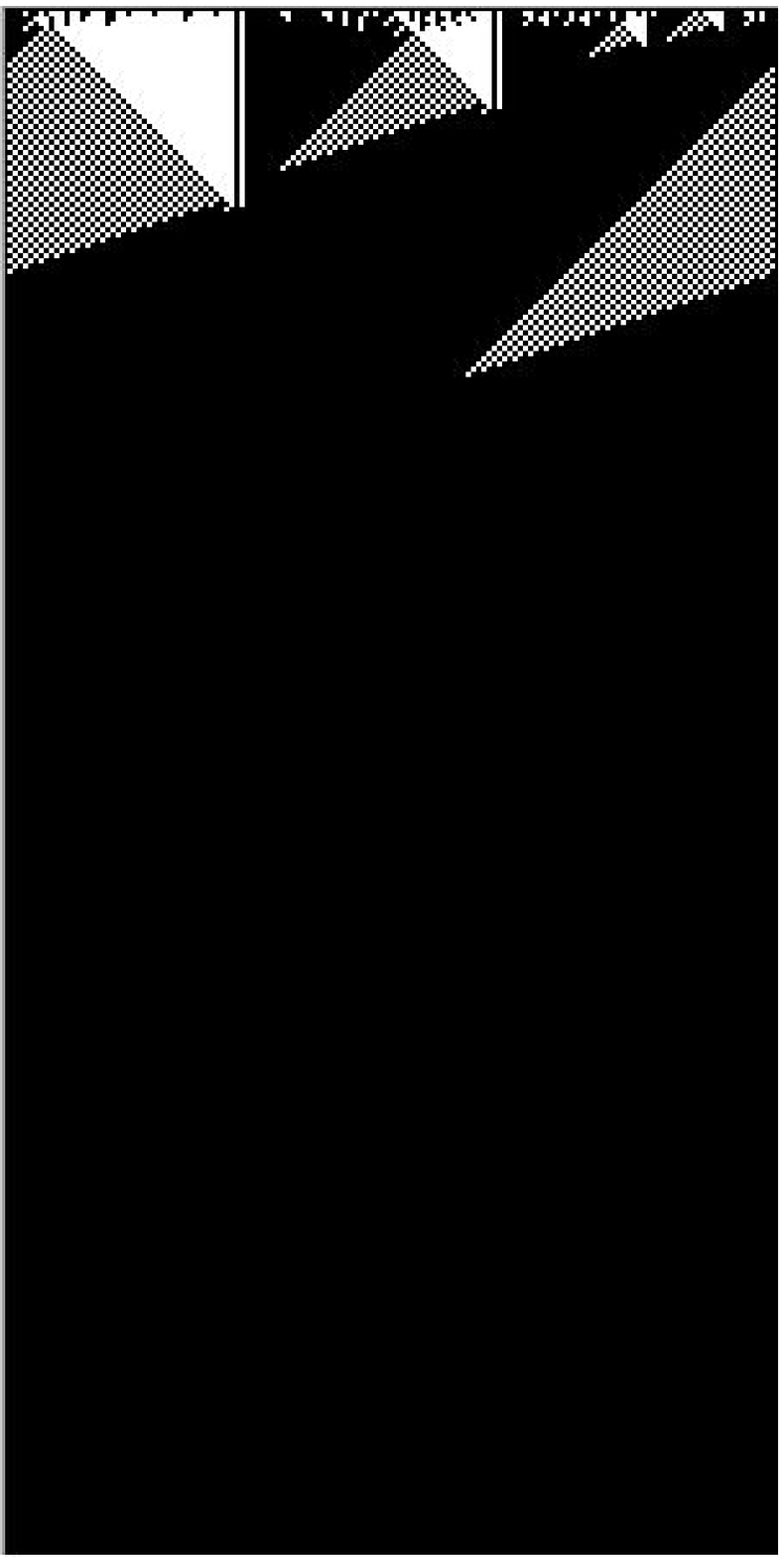,width=4.0cm,height=6cm} }  \protect  \\
(a)   & &  (b)   \\
\end{tabular}
\end{center}

\caption{ Space-time diagram for the GKL rule. The density of
zeros $\rho_0$ is 0.476 in (a) and $\rho_0=0.536$ in (b). State 0
is depicted in white; 1 in black.
\label{sync-ca} }
\end{figure}

CAs are dynamical systems in which space and time are discrete. A
standard CA consists of an array of cells, each of which can be in
one of a finite number of possible states, updated synchronously in discrete time
steps, according to a local, identical transition rule.
Here we will only consider
boolean automata for which the cellular state $s\in\{0,1\}$.
The regular cellular array (grid) is $d$-dimensional, where $d=1,2,3$ is
used in practice. For one-dimensional grids, a cell
is connected to $r$ local neighbors
(cells) on either side where $r$ is  referred to as the \textit{radius}
(thus, each cell has
$2r+1$ neighbors, including itself). 
The \textit{transition rule} contained in each cell is specified in the form of a rule table,
with an entry for every possible neighborhood configuration of states.
The state of a cell at
the next time step is determined by the current states of a surrounding neighborhood
of cells.
Thus, for a linear CA of radius $r$ with
 $ 1 \le r \le N$, the update rule can be written as:
$$
 s^i_{t+1} = \phi(s^{i-r}_t...,s^i_t,...s^{i+r}_t),
$$
\noindent where  $s_i^t$ denotes the state of  site $i$ at  time $t$,
$\phi$ represents the local transition rule,  and  $r$ is  the  CA
radius.\\
The term \textit{configuration} refers to an
assignment of ones and zeros to all the cells at a given time step.
It can be
described by ${\bf s}_{t} =  ( s^0_t,s^1_t,\ldots, s^{N-1}_t)$, where $N$
is  the lattice size. The CAs used here are linear with  periodic boundary
conditions $s^{N+i}_t = s^i_t$ i.e., they are topologically rings.\\
A  global update  rule $\Phi$  can be
defined which applies in parallel to all the cells:
$$
{\bf s}_{t+1} = \Phi ({\bf s}_ t).
$$
\noindent The global map $\Phi$ thus defines the time evolution of the whole
CA.\\
To visualize the behavior of a CA one can use a
two-dimensional space-time diagram, where the horizontal axis depicts the
configuration ${\bf s}_{t}$ at a certain time $t$ and the vertical axis depicts successive time
steps, with time increasing down the page (for example, see figure~\ref{sync-ca}).

\subsection{The Majority Problem}
\label{maj}

The density task is a prototypical distributed
computational problem for CAs. For a finite CA of size $N$ it is defined as follows.
Let $\rho_0$ be the fraction of 1s in the \textit{initial configuration} (IC) ${\bf s}_{0}$.
The task is to determine whether $\rho_0$ is greater than or less than $1/2$. In
this version, the problem is also known as the \textit{majority} problem.
If $\rho_0 > 1/2$ then the CA must
relax to a fixed-point configuration of all 1's that we indicate as
$(1)^N$; otherwise it must relax to a fixed-point
configuration of
all 0's, noted $(0)^N$, after a number of time steps of the order of the grid size
$N$. Here $N$ is set to 149, the value that has been customarily used in research on the
density task (if $N$ is odd one avoids the case $\rho_0=0.5$ for which the problem is undefined).\\
This computation is trivial  for a computer having a central
control. Indeed, just scanning the array and adding up the number
of, say, 1 bits will provide the answer in $O(N)$ time. However, it
is nontrivial for a small radius one-dimensional CA since such a CA can only transfer
information at finite speed relying on local information exclusively, while density
is a global property of the configuration of states \cite{mitchelletal93}.\\
It has been shown
that the density task cannot be solved perfectly by a uniform, two-state CA
with finite radius
\cite{landbelew95}, although a slightly modified version of the task can be shown to
admit perfect solution by such an automaton \cite{capcarrereEtal96}. It can also be 
solved perfectly by a combination of automata \cite{fuks97}.

\subsection{Previous Results on the Majority task}
\label{res}

The lack of a perfect solution does not prevent one from searching for imperfect solutions
of as good a quality as possible. In general, given a desired global behavior for a
CA (e.g., the density task capability), it is extremely difficult to infer the local
CA rule that will give rise to the emergence of the computation sought. This is
because of the possible nonlinearities and large-scale collective effects that
cannot in general be predicted from the sole local CA updating rule, even if it is
deterministic. Since exhaustive evaluation of all possible rules is out of the
question except for elementary ($d=1,r=1$) and perhaps radius-two automata, one possible solution of the
problem consists in using evolutionary algorithms, as first proposed by Packard in
\cite{packard88} and further developed by Mitchell \textit{et al.} in
\cite{mitchelletal94a,mitchelletal93}.\\
The \textit{performance} of the best rules found at the end of the evolution is evaluated on
a larger sample of ICs and it is defined as the fraction of correct classifications
over $n=10^4$ randomly chosen ICs. The ICs are sampled
according to a binomial distribution (i.e., each bit is independently drawn with
probability $1/2$ of being 0). \\
Mitchell and coworkers performed a number of studies on the emergence of
synchronous CA strategies for the density task (with $N=149$) during evolution
\cite{mitchelletal94a,mitchelletal93}.
Their results are significant since they represent one of the few instances where the
dynamics of emergent computation in complex, spatially extended systems can be
understood.
In summary, these findings can be subdivided into those pertaining to
the evolutionary history and those that are part of the ``final'' evolved automata. For
the former, they essentially observed that, in successful evolution experiments, the
fitness of the best rules increases in time according to rapid jumps, giving rise to
what they call ``epochs'' in the evolutionary process. Each epoch corresponds
roughly to a new, increasingly sophisticated solution strategy. Concerning the final
CA produced by evolution, it was noted that, in most runs, the GA found
unsophisticated strategies that consisted in expanding sufficiently large blocks
of adjacent 1s or 0s. This ``block-expanding'' strategy is unsophisticated in that
it mainly uses local information to reach a conclusion. As a consequence, only those
IC that have low or high density are classified correctly since they are more likely
to have extended blocks of 1s or 0s. In fact, these CA have a performance around
$0.6$. However, some of the runs gave solutions that presented novel, more
sophisticated features that yielded better performance (around $0.77$) on a wide
distribution of ICs. However, high-performance
automata have evolved only nine times out of $300$ runs of the genetic
algorithm. 
This clearly shows that it is very difficult for genetic algorithm 
to find good solutions in this the search space.\\
These new strategies rely on traveling signals that transfer
spatial and temporal information about the density in local regions through the
lattice. An example of such a strategy is given in Figure~\ref{sync-ca}, where the
behavior of the so-called GKL rule is depicted \cite{mitchelletal93}.
The GKL rule
is hand-coded  but its behavior is similar to that of the best solutions found
by evolution. 
Crutchfield and coworkers have developed sophisticated methodologies for studying
the transfer of long-range signals and the emergence of computation in evolved CA.
This framework is known as ``computational mechanics'' and it describes the
intrinsic CA computation in terms of regular domains, particles, and particle
interactions. Details can be found in \cite{hansonCrutch95,hordetal98,crutch-mitch-das-03}.\\
Andre \textit{et al.} in \cite{andreetal96b} have been able to artificially
evolve a better CA by using genetic
programming. Finally, Juill\'e and Pollack \cite{juille98} obtained
still better CAs by using a coevolutionary algorithm. Their coevolved CA has performance
about 0.86, which is the best result known to date.

\section{Fitness Landscapes}
\label{fl}

\subsection{Introduction}

First we recall a few fundamental concepts about fitness landscapes (see
\cite{stadler-02,jones95b} for a more detailed treatment).
A landscape is a triplet $(S, f, N)$ where $S$ is a set of
\textit{potential solutions} (also called search space),
 $N: S \rightarrow 2^{S}$, a \textit{neighborhood} structure, is a function that
 assigns to every $s \in S$ a set of neighbors $N(s)$,
and $f:S \mapsto \Real$ is a fitness function that can be pictured as the ``height'' of 
the corresponding potential solutions.\\
Often a topological concept of \textit{distance} $d$ can be associated to a
neighborhood $N$. A distance $d: S \times S \mapsto \Real{^+}$ is a function 
that associates with any two configurations in $S$ a nonnegative
real number that verifies well-known properties.\\
For example, for a binary coded GA, the fitness landscape $S$ is constituted
by the boolean hypercube $B=\{0,1\}^l$ consisting of the $2^l$ solutions
for strings of length $l$ and the associated fitness values. The neighborhood
of a solution, for the one-bit
random mutation operator, is the set of points $y \in B$ that are reachable
from $x$ by flipping one bit. A
natural definition of distance for this landscape is the well-known
\textit{Hamming} distance. 

Based on the neighborhood notion, one can define \textit{local optima} as
being configurations $x$ for which (in the case of maximization):
$ \forall y \in N(x), f(y) \leq f(x)$

Global optima are defined as being the absolute maxima (or minima) in the
whole of $S$. Other features of a landscape such as basins, barriers,
or neutrality can be defined likewise\cite{stadler-02}. Neutrality is a particularly
important notion in our study, and will be dealt with further.

A notion that will be used in the rest of this work is that of a
\textit{walk} on a landscape. A walk $\Gamma$ from $s$ to $s^{'}$ is a sequence
$\Gamma = (s_0, s_1, \ldots, s_m)$ of solutions belonging to $S$ 
where $s_0=s$, $s_m = s^{'}$ and $\forall i \in [1,m]$, $s_i$ is a neighbor of $s_{i-1}$.
The walk can be random, for instance
solutions can be chosen with uniform probability from the neighborhood,
as in random sampling, or according to other weighted non-uniform
distributions, as in
Monte Carlo sampling, for example. It can also be obtained through the
repeated application of a ``move'' operator, either stochastic or
deterministic, defined on the landscape, such as a form of mutation or a
deterministic hill-climbing strategy.

\subsection{Neutrality}
\label{subsec-neutrality}

The notion of neutrality has been suggested by Kimura \cite{KIM:83} in his study of the evolution 
of molecular species. According to this view, most mutations are neutral (their effect on fitness 
is small) or lethal.\\
In the analysis of fitness landscapes, the notion of neutral mutation appears to be useful
\cite{reidys01neutrality}. Let us thus define more precisely the notion of neutrality for
fitness landscapes.

\definition{A {\it test of neutrality} is a predicate $isNeutral : 
S \times S \rightarrow \lbrace true,\linebreak  false  \rbrace$ that assigns to every $(s_1, s_2) \in S^2$ 
the value $true$ if there is a small difference between $f(s_1)$ and $f(s_2)$.}

For example, usually $isNeutral(s_1, s_2)$ is $true$ if $f(s_1) = f(s_2)$. 
In that case, $isNeutral$ is an equivalence relation. 
Other useful cases are $isNeutral(s_1, s_2)$ is $true$
 if $|f(s_1) - f(s_2)| \leq 1/M$ with $M$ is the population size.
When $f$ is stochastic, $isNeutral(s_1, s_2)$ is $true$ if $|f(s_1) - f(s_2)|$ is under the evaluation error.

\definition{For every $s \in S$, the \textit{neutral neighborhood} of $s$ is the set 
$N_{neut}(s) = \lbrace s^{'} \in N(s) ~|~ isNeutral(s,s^{'}) \rbrace$ and the \textit{neutral degree}
 of $s$, noted $nDeg(s)$ is the number of neutral neighbors of $s$, 
 $nDeg(s) = \sharp(N_{neut}(s) - \lbrace s \rbrace$).}\\
A fitness landscape is neutral if there are many solutions with high neutral degree.
In this case, we can imagine fitness landscapes with some plateaus called 
\textit{neutral networks}. 
Informally, 
there is no significant difference of fitness between solutions on neutral networks and the population drifts around on them.

\definition{A \textit{neutral walk} $W_{neut} = (s_0, s_1, \ldots, s_m)$ from $s$ to $s^{'}$ 
is a walk from $s$ to $s^{'}$ where for all $(i,j) \in [ 0, m ]^2$ , $isNeutral(s_i, s_j)$ is $true$.}\\
\definition{A \textit{Neutral Network}, denoted $NN$, is a graph $G=(V,E)$ where the set $V$ of vertices is 
the set of solutions belonging to $S$ such that for all $s$ and $s^{'}$ from $V$ there is a neutral
walk $W_{neut}$ belonging to $V$ from $s$ to $s^{'}$, and two vertices are connected by an edge of $E$ if 
they are neutral neighbors.}\\

\definition{A \textit{portal} in a $NN$ is a solution which has at least one neighbor with greater fitness.}

\subsection{Statistical Measures on Landscapes}
\label{measures}

\subsubsection{Density of States}
\label{dos}

H. Ros\'e \textit{et al.} \cite{rose96} develop the \textit{density of states} approach
(DOS) by plotting the number of sampled solutions in the search space with the same fitness 
value. 
Knowledge of this density allows to evaluate the performance of random search or random initialization
of metaheuristics. DOS gives the probability of having a given fitness value when a
solution is randomly chosen. The tail of the distribution at optimal fitness value gives
a measure of the difficulty of an optimization problem: the faster the decay, the harder the
problem.

\subsubsection{Neutrality}
\label{subsubsec-neutrality}

To study the neutrality of fitness landscapes, we should be able to measure and describe
a few properties of $NN$. The following quantities are useful. The \textit{size} $\sharp NN$ i.e., the
number of vertices in a $NN$,
the \textit{diameter}, which is the longest distance over the distance\footnote{the distance is the shortest length path between two nodes} between two solutions belonging to $NN$. 
The \textit{neutral degree distribution} of solutions
is the degree distribution of the vertices in a $NN$. Together with the size and the diameter, it
gives information which plays a
role in the dynamics of metaheuristic \cite{NIM:99,wilke01}.
Huynen \cite{huynen96exploring} defined the \textit{innovation rate} of $NN$ to explain the advantage
 of neutrality in fitness landscapes. This rate is the number of new,
 previously unencountered fitness values observed in the neighborhood of solutions along a
 neutral walk on $NN$. Finally, $NN$ \textit{percolate} the landscape if they come arbitrarily
 close to almost any every other $NN$ ; this means that, if the innovation rate is high, a
 neutral path could be a good way to explore the landscape.\\
Another way to describe $NN$ is given by the \textit{autocorrelation of neutral degree} along a
neutral random walk \cite{bastolla03}. 
From neutral degree collected along this neutral walk, 
we computed its autocorrelation (see section \ref{acf}). 
The autocorrelation  measures the correlation structure of a $NN$.
If the correlation is low, the variation of neutral degree is low ;
and so, there is some areas in $NN$ of solutions which have nearby neutral degrees. \\

\subsubsection{Fitness Distance Correlation}
\label{fdc}

This statistic was first proposed by Jones \cite{jones95b} with the
aim of measuring the difficulty of problems with a single number.
Jones's approach states that what makes a problem
hard is the relationship between fitness and distance of
the solutions from the optimum. This relationship can be summarized by
calculating the \textit{fitness-distance correlation coefficient (FDC)}.
Given a set $F = \{f_1, f_2, ..., f_m\}$ of $m$ individual fitness values and a
corresponding set $D = \{d_1, d_2, ..., d_m\}$ of the $m$
distances to the nearest global optimum, FDC is defined
as:
$$FDC = \frac{C_{FD}}{\sigma_F \sigma_D}$$
where:
$$C_{FD} = \frac{1}{m} \sum_{i = 1}^{m} (f_i - \overline{f})
(d_i - \overline{d})$$ is the covariance of $F$ and $D$ and
$\sigma_F$, $\sigma_D$, $\overline{f}$ and $\overline{d}$ are the
standard deviations and means of $F$ and $D$. Thus, by definition,
FDC $\in[-1,1]$. As we hope that
fitness increases as distance to a global optimum decreases (for maximization
problems), we expect that, with an ideal fitness function, FDC will
assume the value of $-1$. According to Jones \cite{jones95b}, GA problems can be
classified in three classes, depending on the value of the FDC coefficient:

\begin{itemize}

\item \textit{Misleading} ($FDC \ge 0.15$), in which fitness increases
with distance.

\item \textit{Difficult} ($-0.15 < FDC < 0.15$) in which there is
virtually no correlation between fitness and distance.

\item \textit{Straightforward} ($FDC \le -0.15$) in which fitness increases
as the global optimum approaches.

\end{itemize}

The second class corresponds to problems for which the FDC coefficient does not bring any information. 
The threshold interval $[-0.15,0.15]$ has been empirically determined by Jones. 
When FDC does not give a clear indication i.e.,
in the interval $[-0.15,0.15]$, examining the scatterplot of fitness versus distance can be useful.

The FDC has been criticized on the grounds that counterexamples can
be constructed for which the measure gives wrong results \cite{alt97,quick98,Clergue:2002:GFBbCoTF}.
Another drawback of FDC is the fact that it is not a \textit{predictive}
measure since it requires knowledge of the optima. Despite its
shortcomings, we use FDC here as another way of characterizing problem difficulty
because we know some optima
and we predict whether or not it is easy to reach those local optima.

\subsubsection{The Autocorrelation Function and the Box-Jenkins approach}
\label{acf}

Weinberger \cite{WEI:91,WEI:90} introduced the {\it autocorrelation function} and 
the {\it correlation length} of random walks to measure the correlation structure of 
fitness landscapes. Given a random walk $( s_t, s_{t+1}, \ldots  )$,
the autocorrelation function $\rho$ of a fitness function $f$ is the autocorrelation
function of time series $( f(s_t), f(s_{t+1}), \ldots )$ :
$$
\rho(k) = \frac{E[f(s_t) f(s_{t+k})] - E[f(s_t)]E[f(s_{t+k})]}{var(f(s_t))}
$$
where $E[f(s_t)]$ and $var(f(s_t))$ are the expected value and the variance of $f(s_t)$.
Estimates $r(k)$ of autocorrelation coefficients $\rho(k)$ can be calculated with a time
series $( s_1, s_2, \ldots, s_{L} )$ of length $L$ :
$$
r(k) = \frac{\sum_{j=1}^{L-k}(f(s_j) - \bar{f}) (f(s_{j+k}) - \bar{f})}{\sum_{j=1}^{L}(f(s_j) - \bar{f})^2}
$$
where $\bar{f} = \frac{1}{L} \sum_{j=1}^{L} f(s_j)$, and $L >> 0$.
A random walk is representative of the entire landscape when the landscape
is statistically isotropic. In this case, whatever the starting point of random walks and the selected
neighbors during the walks, estimates of $r(n)$ must be nearly the same. Estimation error diminishes with
the walk length.

The correlation length $\tau$ measures how the autocorrelation function decreases and it
summarizes the ruggedness of the landscape: the larger the correlation length, the smoother is the
landscape.
Weinberger's definition $\tau = - \frac{1}{ln(\rho(1))}$ makes the assumption that the autocorrelation function
decreases exponentially. Here we will use another definition that comes from a more general analysis
of time series, the Box-Jenkins approach \cite{boxJenkins70}, introduced in the field of fitness
landscapes by Hordijk \cite{hordijk96measure}.
The time series of fitness values will be approached by an {\it autoregressive moving-average} (ARMA) model.
In ARMA$(p,q)$ model, the current value depends linearly on the $p$ previous values and the $q$ previous 
white noises.
$$
f(s_t) = c + \sum_{i=1}^{p} \alpha_i f(s_{t-i}) + \epsilon_{t} + \sum_{i=1}^{q} \beta_i \epsilon_{t-i}
\ \mathrm{where}\ \epsilon_t \ \mathrm{are\ white\ noises.}
$$

The approach consists in the iteration of three stages \cite{boxJenkins70}.
The \textit{identification stage} consists in determining the value of $p$ and $q$ using the autocorrelation 
function (acf) and the partial autocorrelation function (pacf) of the time series. The \textit{estimation
stage} consists in determining the values $c$, $\alpha_i$ and $\beta_i$ using the pacf. The significance
 of this values is tested by a t-test. The value is not significant if t-test is below $2$. The
 \textit{diagnostic checking stage} is composed of two parts. The first one checks the adequation between data and 
 estimated data. We use the {\it square correlation $R^2$} 
 between observed data of the time series and estimated data produced by the model and the {\it Akaide information criterion AIC}:
$$AIC(p,q) = log(\hat{\sigma}^2) + 2 (p+q) / L \ \mathrm{where}\ \hat{\sigma}^2 = L^{-1}\sum_{j=1}^{L}(y_j - \hat{y}_j)^2$$
The second one checks the white noise of residuals which is the difference between observed data value 
and estimated values. For this, the autocorrelation of residuals and the p-value of Ljung-Box test are computed.

\subsection{Fitness Cloud and NSC}
\label{fc}

We use the \textit{fitness cloud} (FC) standpoint, first introduced in \cite{SEB:03} by V\'erel and coworkers. 
The fitness cloud relative to the local search operator $op$ is the conditional bivariate probability density $P_{op}(Y = \tilde{\varphi} ~|~ X = \varphi)$ of reaching a solution of fitness value $\tilde{\varphi}$ from a solution of fitness value $\varphi$ applying the operator $op$. 
To visualize the fitness cloud in two dimensions,
we plot the scatterplot $FC = \lbrace ( \varphi , \tilde{\varphi} ) ~|~ P_{op}(\varphi, \tilde{\varphi} ) \not= 0 \rbrace$.

In general, the size of the search space does not allow
to consider all the possible individuals, when trying to draw
a fitness cloud.
Thus, we need to use samples to estimate it.
We prefer to sample
the space according to a distribution that gives more weight to
``important'' values in the space, for instance those at a higher 
fitness level.
This is also the case of any biased searcher
such as an evolutionary algorithm,
simulated annealing and other heuristics, and thus 
this kind of sampling process
more closely simulates the way in which the program space would be 
traversed by
a searcher. 
So, we use the Metropolis-Hastings technique \cite{monte-carlo}
to sample the search space.


The Metropolis-Hastings sampling technique is
an extension of the Metropolis algorithm to non-symmetric
stationary probability distributions.
It can be defined as follows. 
Let $\alpha$ be the function defined as:

$$ \alpha(x, y) = min\{1, \frac{y}{x}\}, $$

\

\noindent
and $f(\gamma_{k})$ be the fitness of individual $\gamma_{k}$. A sample
of individuals $\{\gamma_1, \gamma_2, \ldots, \gamma_{n}\}$ is built
with the algorithm shown in figure \ref{mehasalgo}.

\begin{figure}[!ht]
\vspace*{0.5cm}
\begin{small}
\begin{tt}
\fbox{ \parbox{12cm}{
\noindent
{\bf begin} \\
\hspace*{0.5cm}$\gamma_1$ is generated uniformly at random; \\
\hspace*{0.5cm}{\bf for} $k := 2$ {\bf to} $n$ {\bf do} \\
\hspace*{1cm}1. an individual $\delta$ is generated uniformly at random; \\
\hspace*{1cm}2. a random number $u$ is generated from a \\ 
\hspace*{1.4cm}   uniform $(0,1)$ distribution; \\
\hspace*{1cm}3. {\bf if} $(u \le \alpha (f(\gamma_{k-1}), f(\delta)))$ \\
\hspace*{2cm}{\bf then} $\gamma_{k} := \delta$ \\
\hspace*{2cm}{\bf else} {\bf goto} 1. \\
\hspace*{1.7cm}{\bf endif} \\
\hspace*{1cm}4. $k := k+1$; \\
\hspace*{0.5cm}{\bf endfor} \\
{\bf end}
}}
\end{tt}
\end{small}
\vspace{-0.2cm}
\caption{The algorithm for sampling a search space with
the Metropolis-Hastings technique.}
\label{mehasalgo}
\end{figure}


In order to algebraically extract some information from the fitness
cloud, in \cite{gecco-04,tesi}, we have defined a measure, called
{\it negative slope coefficient} ({\it nsc}).
The abscissas of a scatterplot can be
partitioned into $m$ segments $\{I_1, I_2, \ldots, I_m\}$
with various techniques.
Analogously, a partition of the ordinates $\{J_1, J_2, \ldots, \linebreak J_m\}$
can be done, where each segment $J_i$ contains all the ordinates
corresponding to the abscissas contained in $I_i$.
Let $M_1, M_2, \ldots, M_m$ be the averages of the abscissa values
contained inside the segments $I_1, I_2, \ldots, I_m$
and let $N_1, N_2, \ldots, N_m$ be the averages of the ordinate values in 
$J_1, J_2, \ldots, J_m$.
Then we can define the set of segments $\{S_1, S_2, \ldots, S_{m-1}\}$,
where each $S_i$ connects the point $(M_i, N_i)$ to the point
$(M_{i+1}, N_{i+1})$.
For each one of these segments $S_i$, the 
{\it slope} $P_i$ can be calculated as follows:
$$ P_i = \frac{N_{i+1} - N_i}{M_{i+1} - M_{i}} $$
\noindent
Finally, we can define the NSC as:
$$ nsc = \displaystyle \sum_{i = 1}^{m-1} c_i ,   \ \ \mbox{where:}  \  \forall i \in [1, m) \ \ c_i = min (P_i, 0)  $$

\noindent
We hypothesize that {\it nsc} can give some indication of problem 
difficulty in the following
sense: if {\it nsc}$=0$, the problem is easy, if {\it nsc}$<0$ 
the problem is difficult and the value
of {\it nsc} quantifies this difficulty: the smaller its value,
the more difficult
the problem.
In other words, according to our hypothesis, a problem is difficult if at least
one of the segments $S_1, S_2, \ldots, S_{m-1}$ has a negative slope and
the sum of all the negative slopes gives a measure of problem hardness.
The idea is that the presence of a segment with negative slope
indicates a bad evolvability for individuals having
fitness values contained in that segment.

\section{Analysis of the Majority Problem Fitness Landscape}
\label{sec-maj-land}

\subsection{Definition of the fitness landscape}
\label{subsec-maj-land}

As in Mitchell \cite{mitchelletal93}, we use CA of radius $r=3$ and configurations of length $N=149$.
The set $S$ of potential solutions of the Majority Fitness Landscape is the set of binary string which represent the possible CA rules. 
The size of $S$ is $2^{2^{2r+1}} = 2^{128}$, and each automaton should be tested on the $2^{149}$ possible different ICs.
This gives $2^{277}$ possibilities, a size far too large to be searched exhaustively.
Since performance can be defined in several ways, the consequence is that for each feasible CA in
the search space, the associated fitness can be different, and thus effectively inducing
different landscapes.
In this work we will use one type of performance measure based on the fraction of $n$ initial
configurations that are correctly classified from one sample. 
We call \textit{standard performance} (see also section \ref{res}) the performance when the sample is drawn from a binomial 
distribution (i.e., each bit is independently drawn with probability $1/2$ of being 0).
Standard performance is a hard measure
because of the predominance in the sample of ICs close to $0.5$ and
it has been typically employed to measure a CA's capability on the density task.

\begin{figure}[!ht]
\begin{center}
\mbox{
  \epsfig{figure=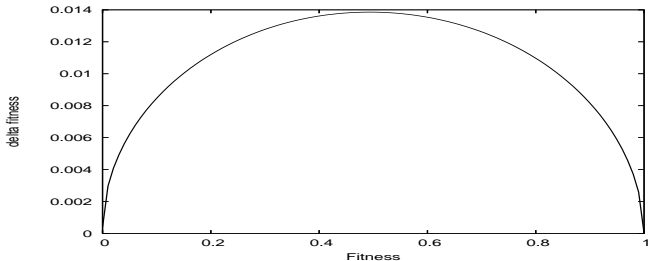,width=9cm,height=3.5cm} } 
\end{center}
\caption{Error of standard performance as a function of standard performance given by t-test with $95$ percent of confidence with sample of size $n=10^4$. 
Bitflip is neutral if the absolute value of the difference between the two fitnesses is above the curve.
\label{fig-deltaf} }
\end{figure}

The standard performance cannot be known perfectly due to random variation of samples of ICs. 
The fitness function of the landscape is stochatic one
which allows population of solutions to drift arround neutral networks.
The error of evaluation leads us to define the neutrality of landscape. 
The ICs are chosen independently, so the fitness value $f$ of a solution follows
a normal law $\mathcal{N}(f, \frac{\sigma(f)}{\sqrt{n}})$, where $\sigma$ is the standard deviation
 of sample of fitness $f$, and $n$ is the sample size. 
For binomial sample, $\sigma^2(f) = f (1-f)$, the variance of Bernouilli trial.
Then two neighbors $s$ and $s^{'}$ are neutral neighbors ($isNeutral(s,s^{'})$ is $true$) if a t-test accepts
the hypothesis of equality of $f(s)$ and $f(s^{'})$ with $95$ percent of confidence (fig. \ref{fig-deltaf}).
The maximum number of fitness values statistically different for standard performance
 is $113$ for $n=10^4$, $36$ for $n=10^3$ and $12$ for $n=10^2$.
\\

\subsection{First statistical Measures}
\label{subsec-majLand-measures}

\begin{figure}[!ht]
\begin{center}
\begin{tabular}{ccc}
\mbox{
  \epsfig{figure=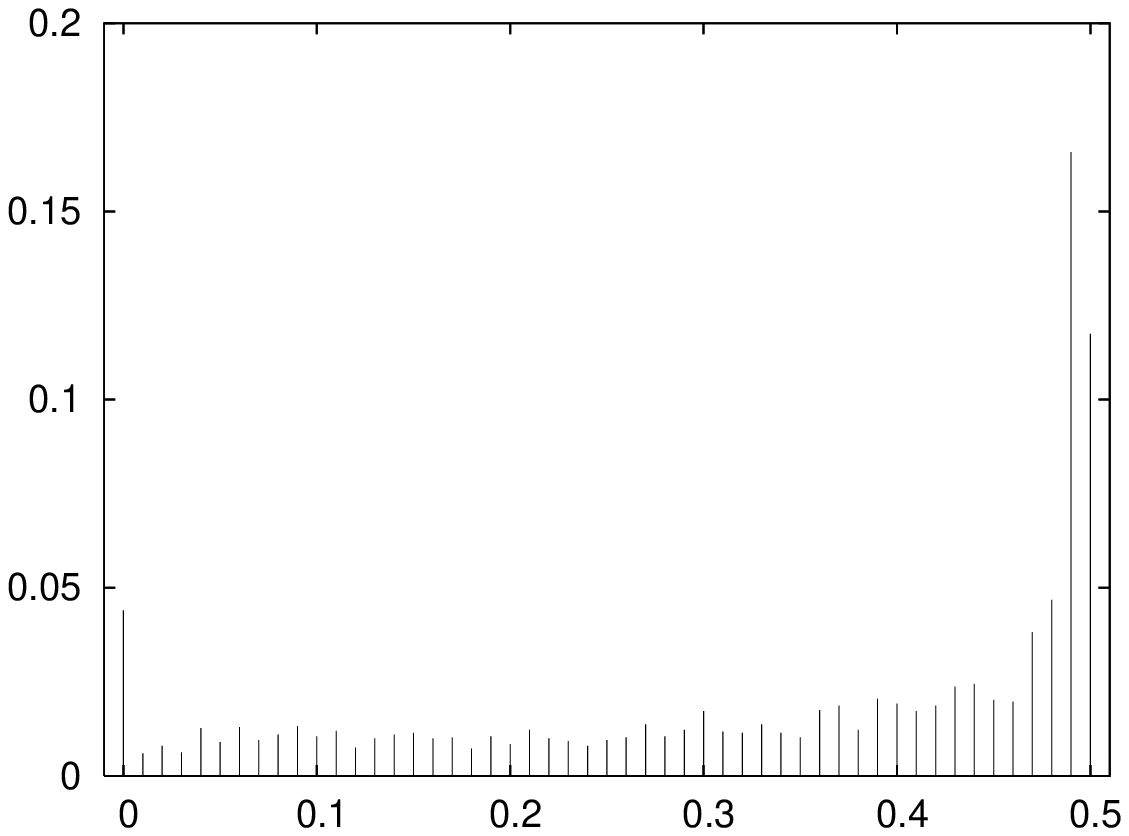,width=6.0cm,height=5.3cm} } & &
\mbox{
  \epsfig{figure=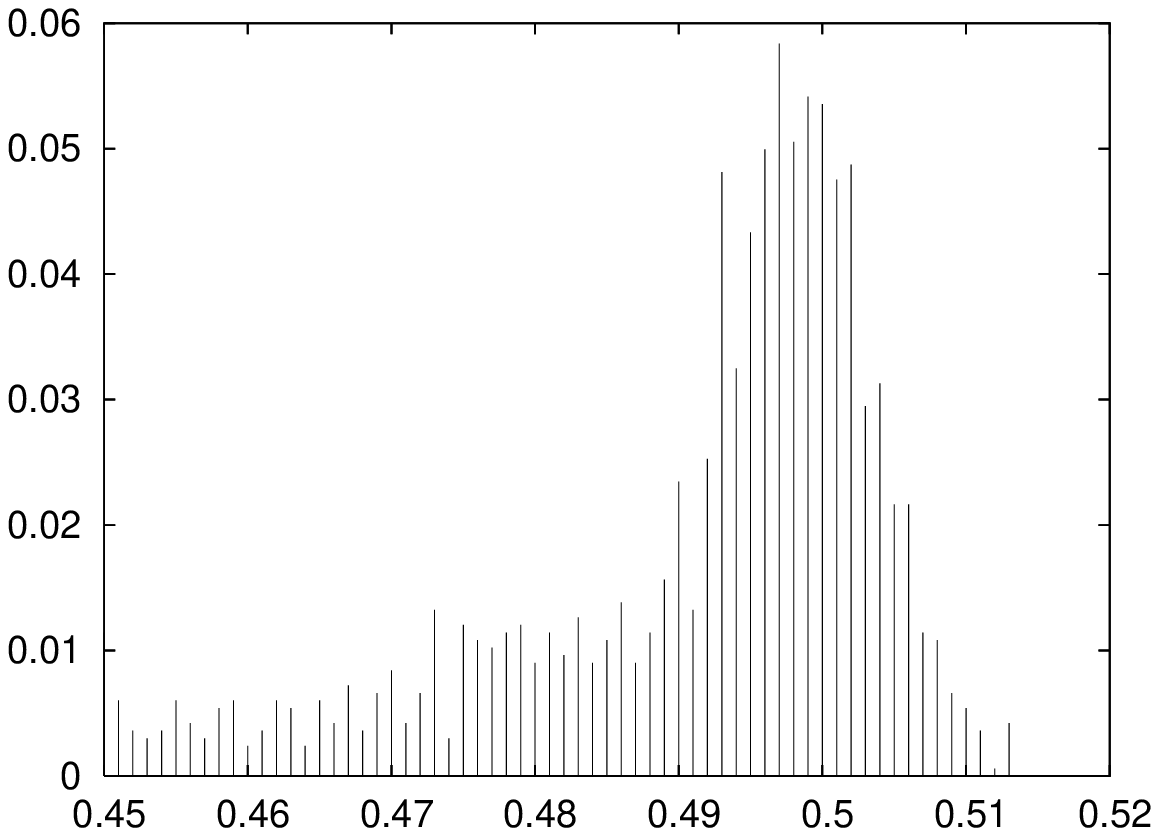,width=6.0cm,height=5.3cm} }  \protect  \\
(a)   & &  (b)   \\
\end{tabular}
\end{center}
\caption{ DOS obtained using the Metropolis-Hastings technique to sample the search space,
	  between fitness values $0.0$ and $0.52$ in (a) and between $0.45$ and $0.52$ in (b).
	  x-axis: fitness values; y-axis: fitness values frequencies.
\label{fig-dos} }
\end{figure}

The DOS of the Majority problem landscape was computed using the uniform random sampling technique. 
The number of sampled points is $4000$ and, among them,
the number of solutions with a fitness value equal to 0 is $3979$.
Clearly, the space appears to be a difficult one to search since the tail of the distribution to the right is non-existent.
Figure \ref{fig-dos} shows the DOS obtained using the Metropolis-Hastings technique. 
This time, over the $4000$ solutions sampled, only $176$
have a fitness equal to zero, and the DOS clearly shows a more
uniform distribution of rules over many different fitness values.

It is important to remark a considerable number of solutions
sampled with a fitness approximately equal to $0.5$.
Furthermore, no individual with a fitness value superior to $0.514$
has been sampled.
For the details of the techniques used to sample the space,
see \cite{gecco-04,tesi}

The autocorrelation along random walks is not significant due to the large number of zero fitness
points and is thus not reported here.

The FDC, calculated over a sample of 4000 individuals generated using the
Metropolis-Hastings technique, are shown in table \ref{fdc-whole}. 
Each value has been obtained using one of the best local optima known to date (see section \ref{best-opt}).
FDC values are approximately close to zero for DAS optimum.
For ABK optimum, FDC value is near
to -0.15, value identified by Jones as the threshold between difficult
and straightforward problems. For all the other optima, FDC are close to $-0.10$.
So, the FDC does not provide information about problem difficulty.

\vspace*{0.3cm}

\begin{table}[!ht]
\caption{FDC values for the six best optima known, calculated over a sample
of 4000 solutions generated with the Metropolis-Hastings sampling technique.}
\label{fdc-whole}
\begin{center}
\begin{tabular}{crrrrrr}
\hline
Rules &
GLK \cite{gacsEtal78}            &
Davis \cite{andreetal96b}        &
Das \cite{dasetal94}             &
ABK \cite{andreetal96b}          &
Coe1 \cite{juille98coevolving}   &
Coe2 \cite{juille98coevolving}   \\
FDC &
 -0.1072 &
 -0.0809 &
 -0.0112 &
 -0.1448 &
 -0.1076 &
 -0.1105 \\
\hline
\end{tabular}
\end{center}
\end{table}

\vspace*{0.3cm}

\begin{figure}[!ht]
\begin{center}
\begin{tabular}{cc}
\mbox{
  \epsfig{figure=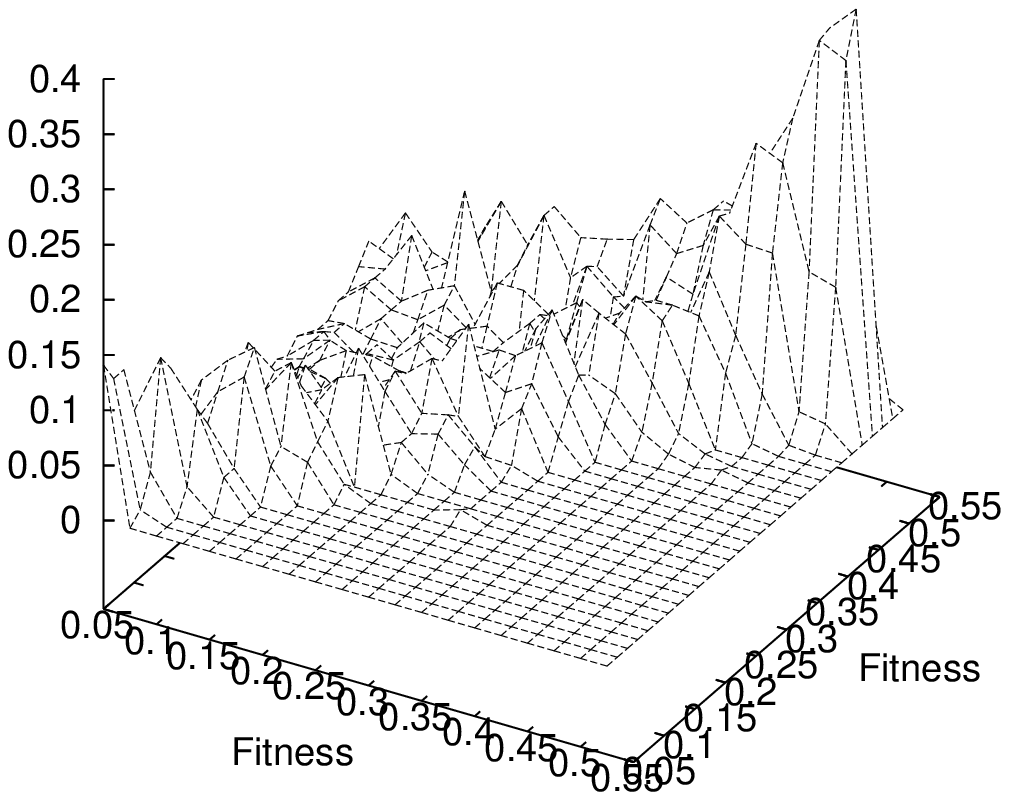,width=6cm,height=6cm} } &
\mbox{
  \epsfig{figure=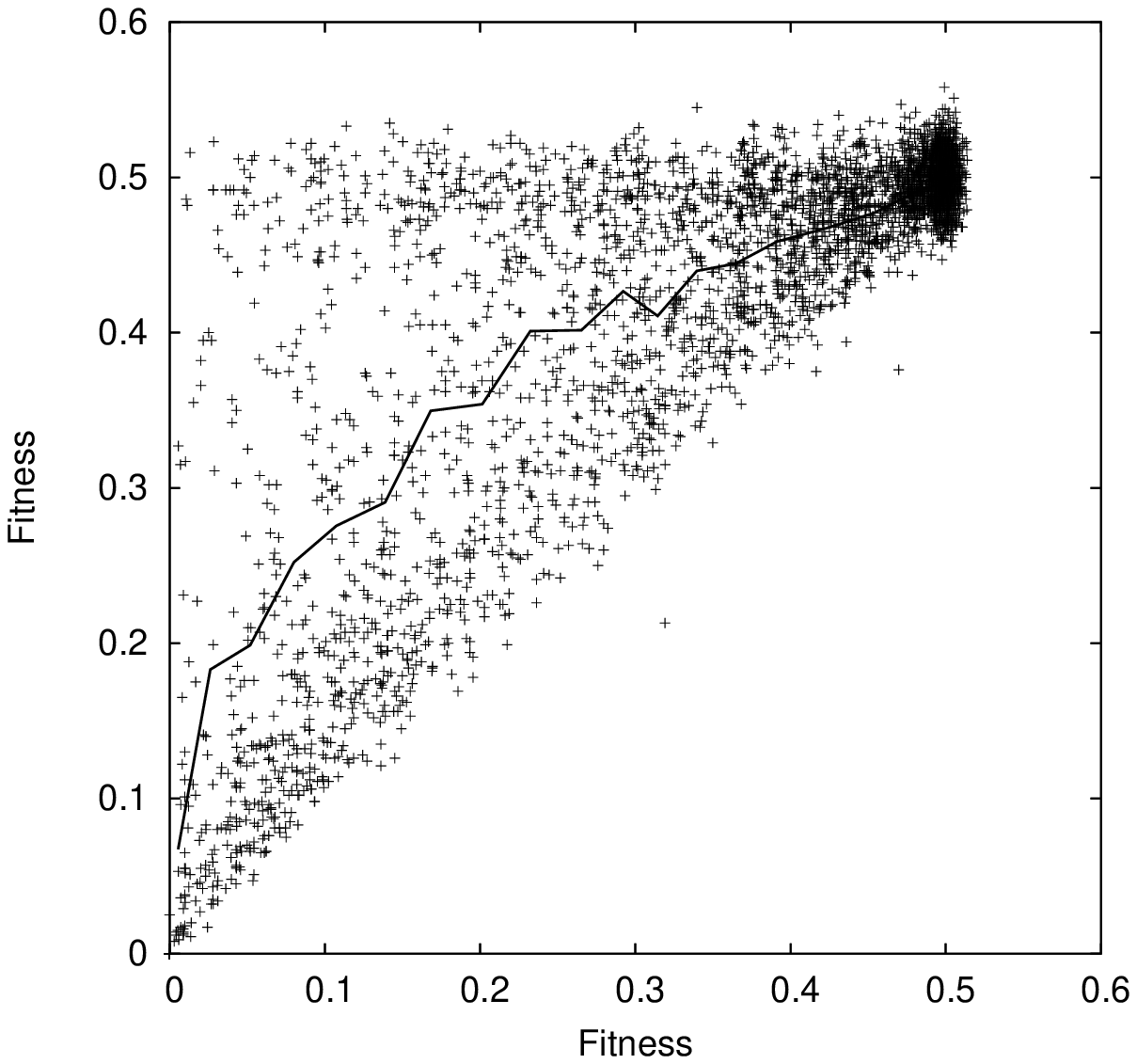,width=6cm,height=6cm} } 
\end{tabular}
\end{center}
\caption{Fitness Cloud and Scatterplot used to calculate the NSC value.
Metropolis-Hastings technique has been used to sample the search space.
\label{nsc-whole} }
\end{figure}

Figure \ref{nsc-whole} shows the fitness cloud, and
the set of segments
used to calculate the NSC.
As this figure clearly shows, the Metropolis-Hastings technique allows to
sample a significant number of solutions with a fitness value higher than zero. 
The value of the NSC for this problem is $-0.7133$, indicating that 
it seems difficult for a local search heuristic to reach fitness values close to $0.5$,
and going further seems to be much harder.

\subsection{Neutrality}
\label{gl-neut}

Computational costs do not allow us to analyze many neutral networks.
In this section we analyze two important large neutral networks ($NN$). 
A large number of CAs solve the majority density problem on only half of ICs because
they converge nearly always on the final configuration $(O)^N$ or $(1)^N$ and thus have performance
about $0.5$. Mitchell \cite{mitchelletal94a}
calls these ``default strategies'' and notices that they are the first stage in the evolution of the
population before jumping to higher performance values associated to ``block-expanding'' strategies
(see section \ref{res}). We will study this large $NN$, denoted $NN_{0.5}$ around standard
performance $0.5$ to
understand the link between $NN$ properties and GA evolution. The other $NN$, denoted $NN_{0.76}$,
is the $NN$ around fitness $0.7645$ which contains one neighbor of a CA found by Mitchell \textit{et al}. 
The description of this ``high'' $NN$ could give clues as how to ``escape'' from $NN$ toward even
higher fitness values.

\begin{table}
\caption{Description of the starting of neutral walks.\label{tab-startNeutralWalk}}

\begin{center}
\begin{tabular}{|l|l|}
\hline
  &
{\scriptsize
00000000   00000110   00010000   00010100   00001010   01011000   01111100   01001101 }\\
0.5004 & 
{\scriptsize
01000011   11101101   10111111   01000111   01010001   00011111   11111101   01010111 }\\

\hline
  &
{\scriptsize
00000101   00000100   00000101   10100111   00000101   00000000   00001111   01110111 }\\
0.7645 & 
{\scriptsize
00000011   01110111   01010101   10000011   01111011   11111111   10110111   01111111 }\\

\hline
\end{tabular}
\end{center}
\end{table}

In our experiments, we perform $5$ neutral walks on $NN_{0.5}$ and $19$ on $NN_{0.76}$. 
Each neutral walk has the same starting point on each $NN$. 
The solution with performance $0.5$ is randomly solution and
the solution with performance $0.76$ is a neighboring solution of solution find by Mitchell (see tab \ref{tab-startNeutralWalk}).
We try to explore the $NN$ by strictly increasing the Hamming
distance from the starting solution at each step of the walk. 
The neutral walk stops when there is no neutral step that increases distance. 
The maximum length of walk is thus $128$.
On average, the length of neutral walks on $NN_{0.5}$ is $108.2$ and $33.1$ on $NN_{0.76}$. 
The diameter (see section \ref{subsubsec-neutrality}) of $NN_{0.5}$ should probably be larger than the one of $NN_{0.76}$.

\begin{figure}[!ht]
\begin{center}
\begin{tabular}{ccc}
\mbox{
  \epsfig{figure=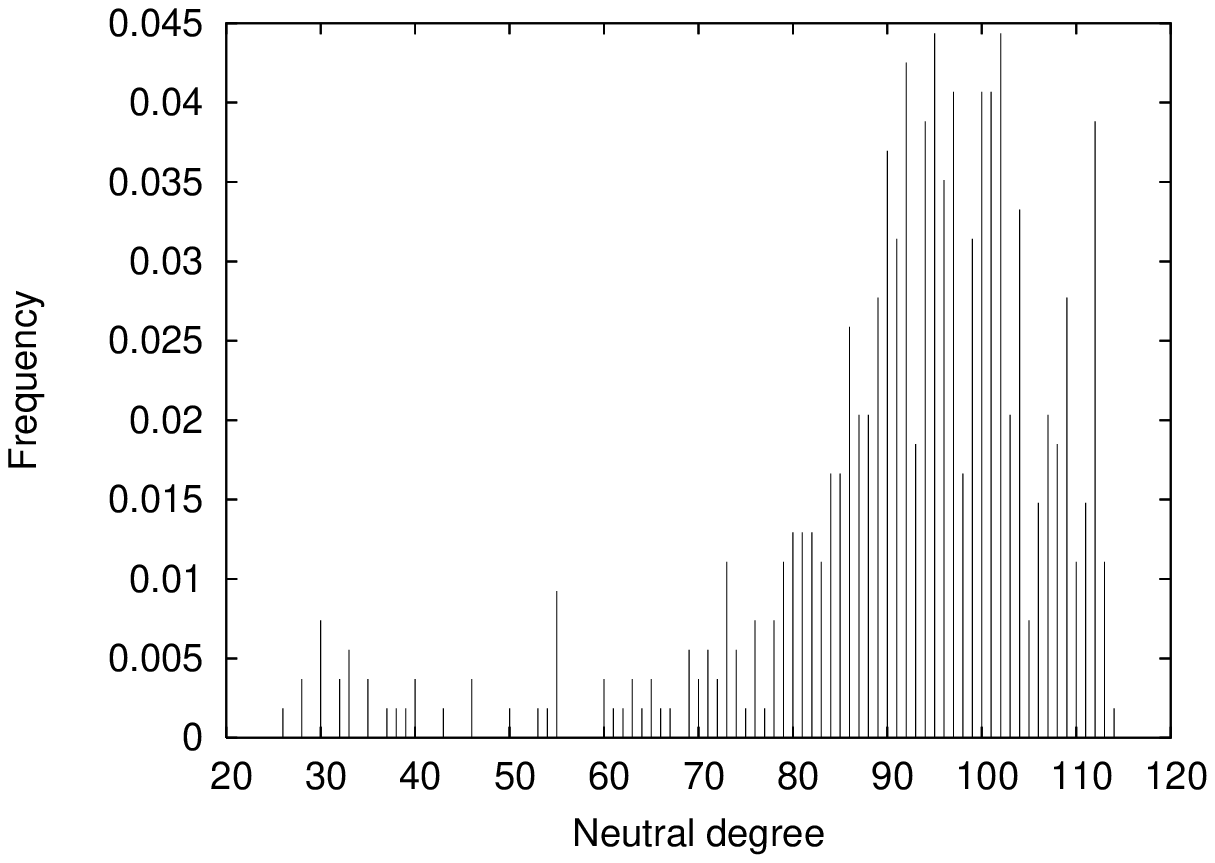,width=6.0cm,height=6cm} } & &
\mbox{
  \epsfig{figure=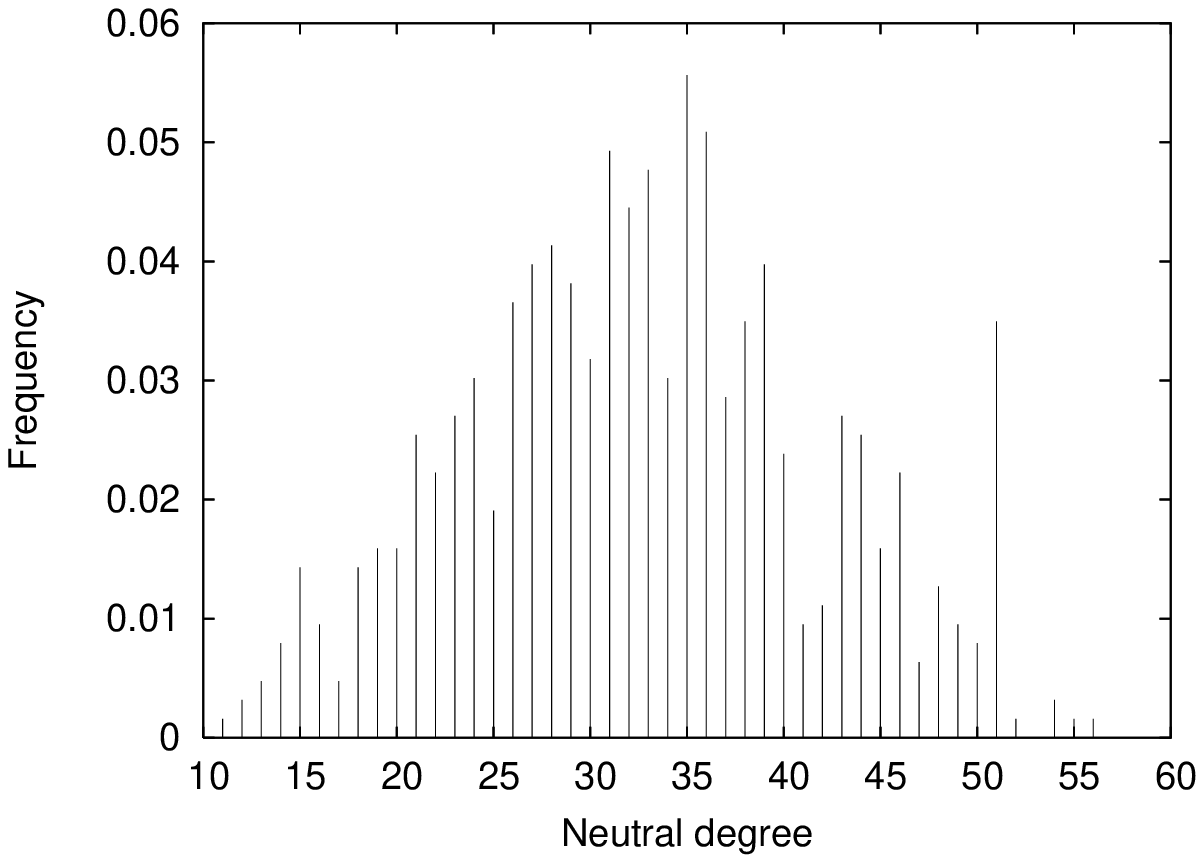,width=6.0cm,height=6cm} }  \protect  \\
(a)   & &  (b)   \\
\end{tabular}
\end{center}
\caption{Distribution of Neutral Degree along all neutral walks on $NN_{0.5}$ in (a) and $NN_{0.76}$ in (b).
\label{fig-nDeg_distri} }
\end{figure}

Figure \ref{fig-nDeg_distri} shows the distribution of neutral degree collected along all neutral walks.
The distribution is close to normal for $NN_{0.76}$. For $NN_{0.5}$ the distribution
is skewed and approximately bimodal with a strong peak around $100$ and a small peak around $32$. 
The average of neutral degree on
$NN_{0.5}$ is $91.6$ and standard deviation is $16.6$; on $NN_{0.76}$, the average is $32.7$ and 
the standard deviation is $9.2$.
The neutral degree for $NN_{0.5}$ is very high : $71.6$ \% of neighbors are neutral neighbors.
For $NN_{0.76}$, there is $25.5$ \% of neutral neighbors. It can be compared to the average 
neutral degree overall neutral $NKq$-landscape with $N=64$, $K=2$ and $q=2$ which is $33.3$ \% \cite{Verel:ECAI2004}. 

\begin{figure}[!ht]
\begin{center}
\begin{tabular}{ccc}
\mbox{
  \epsfig{figure=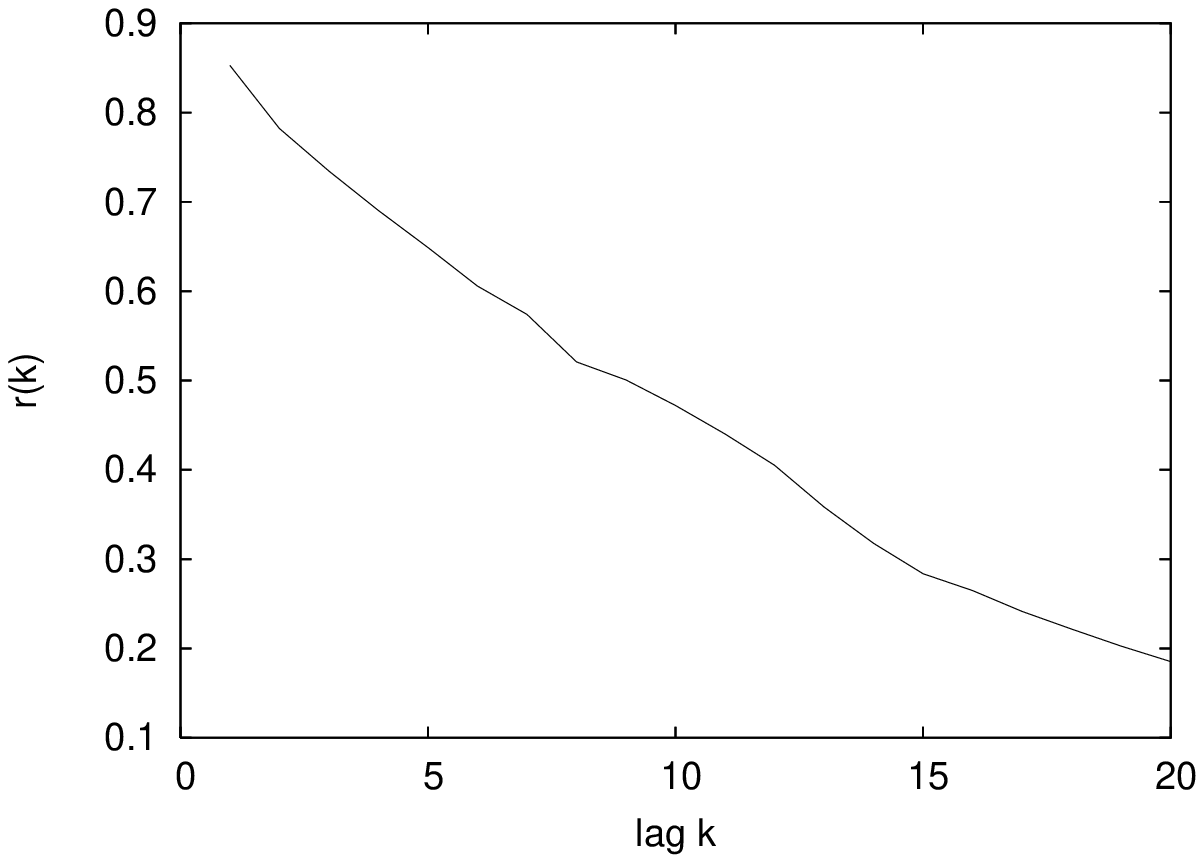,width=6.0cm,height=6cm} } & &
\mbox{
  \epsfig{figure=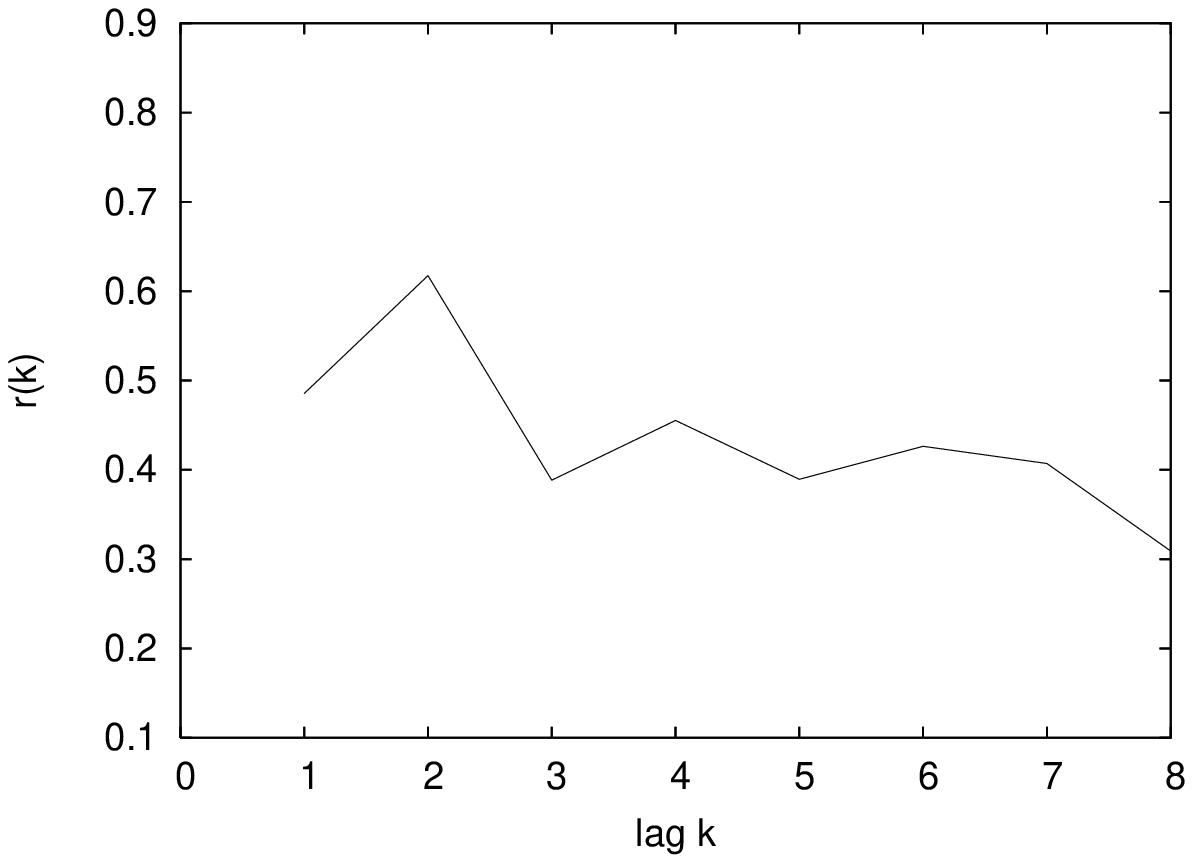,width=6.0cm,height=6cm} }  \protect  \\
(a)   & &  (b)   \\
\end{tabular}
\end{center}
\caption{Estimation of the autocorrelation function of neutral degrees along neutral random walks for 
$NN_{0.5}$ (a) and for $NN_{0.76}$ (b).
\label{fig-nDeg_acf} }
\end{figure}

Figure \ref{fig-nDeg_acf} gives an estimation of the autocorrelation function of neutral
degree of the neutral networks. The autocorrelation function is computed for each neutral walk 
and the estimation $r(k)$ of $\rho(k)$ is given by the average of $r_i(k)$ over all autocorrelation functions. 
For both $NN$, there is correlation. The correlation is higher for $NN_{0.5}$ ($\rho(1)=0.85$) than  for $NN_{0.76}$ ($\rho(1)=0.49$).
From the autocorrelation of the neutral degree, one can conclude that the neutral network
topology is not completely random, since otherwise correlation should have been nearly equal to zero. 
Moreover, the variation of neutral degree is smooth on $NN$; in other words, the neighbors in $NN$ have nearby neutral degrees. So, there is some area where the neutral degree is homogeneous.

\begin{figure}[!ht]
\begin{center}
\begin{tabular}{ccc}
\mbox{
  \epsfig{figure=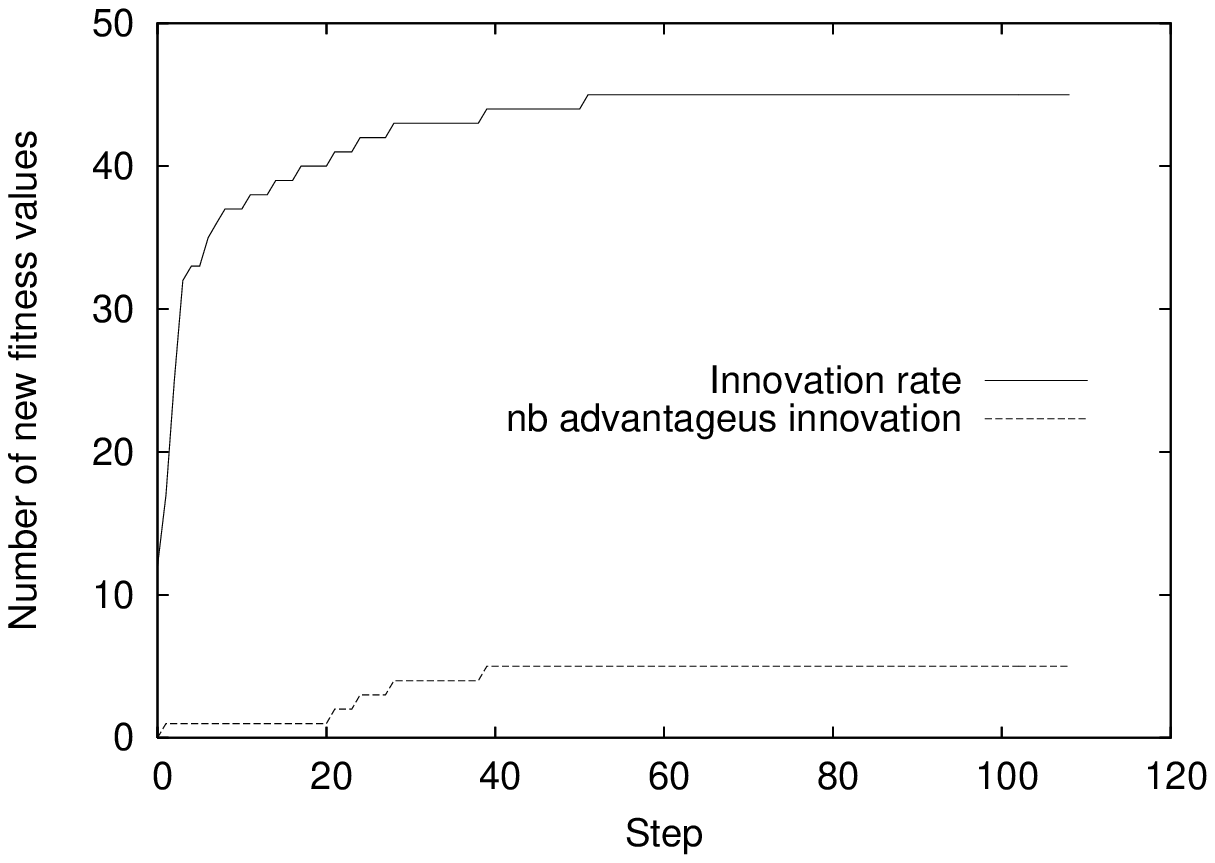,width=6.5cm,height=6cm} } & &
\mbox{
  \epsfig{figure=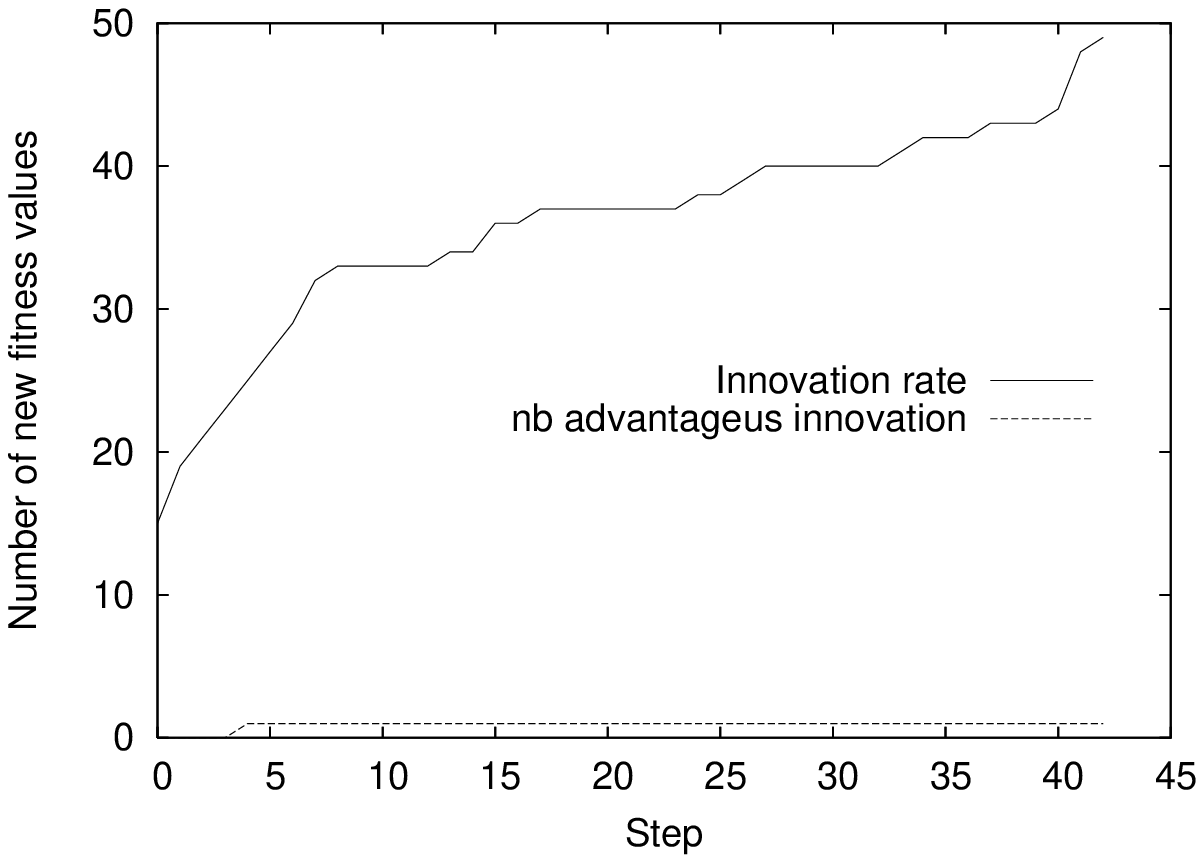,width=6.5cm,height=6cm} }  \protect  \\
(a)   & &  (b)   \\
\end{tabular}
\end{center}
\caption{Innovation rate along one neutral random walk. Fitness of the neutral networks
is $0.5004$ (a) and $0.7645$ (b). \label{fig-innov} }
\end{figure}

The innovation rate and the number of new better fitnesses found 
along the longest neutral random walk for each $NN$ is given in figure \ref{fig-innov}.
The majority of new fitness value found along random walk is deleterious, very few solutions are fitter. 

This study give us a better description of Majority fitness landscape neutrality which have important
consequence on metaheuristic design.
The neutral degree is high. Therefore, the selection operator should take into account the case of equality of
fitness values. Likewise the mutation rate and population size should fit to this neutral degree in order to
find rare good solutions outside $NN$ \cite{barnett01netcrawling}.
For two potential solutions $x$ and $y$ on $NN$, the probability $p$ that at least one solution escaped
from $NN$ is $P(x \not\in NN \cup y \not\in NN) = P(x \not\in NN) + P(y \not\in NN) - P(x \not\in NN \cap y \not\in NN)$. 
This probability is higher when solutions $x$ and $y$ are far due to the correlation of neutral degree in $NN$. 
To maximize the probability of escaping $NN$ the distance between potential solutions of population should be as far as possible on $NN$.
The population of an evolutionary algorithm should spread over $NN$.

\subsection{Study of the best local optima known}
\label{best-opt}

We have seen that it is difficult to have some relevant informations on the Majority Problem 
landscape by random sampling  due to the large number of solutions with zero fitness. In
this section, we will study the landscape from the top.
Several authors have found fairly good solutions for the density problem, either by hand or, especially, 
using evolutionary
algorithms \cite{gacsEtal78,dasetal94,andreetal96b,juille98coevolving}. We will 
consider the six Best Local Optima Known\footnote{In section \ref{gl-profilOpt}, we will show that these are really local optima}, that we call \textit{blok}, with a standard performance
over $0.81$ (tab. \ref{tab-opt6}). In the following, we will see where the \textit{blok} are located and what is the structure of the landscape around these optima.

\begin{table}
\caption{Description and standard performance of the 6 previously known best rules (\textit{blok}) computed on sample size of $10^4$.\label{tab-opt6}}

\begin{center}
\begin{tabular}{|l|l|}
\hline
GLK  &
{\scriptsize
00000000   01011111   00000000   01011111   00000000   01011111   00000000   01011111 }\\
0.815 & 
{\scriptsize
00000000   01011111   11111111   01011111   00000000   01011111   11111111   01011111 }\\

\hline
Das  &
{\scriptsize
00000000   00101111   00000011   01011111   00000000   00011111   11001111   00011111 }\\
0.823 &
{\scriptsize
00000000   00101111   11111100   01011111   00000000   00011111   11111111   00011111 }\\

\hline
Davis  &
{\scriptsize
00000111   00000000   00000111   11111111   00001111   00000000   00001111   11111111 }\\
0.818 &
{\scriptsize
00001111   00000000   00000111   11111111   00001111   00110001   00001111   11111111 }\\

\hline
ABK  &
{\scriptsize
00000101   00000000   01010101   00000101   00000101   00000000   01010101   00000101 }\\
0.824 &
{\scriptsize
01010101   11111111   01010101   11111111   01010101   11111111   01010101   11111111 }\\

\hline
Coe1  &
{\scriptsize
00000001   00010100   00110000   11010111   00010001   00001111   00111001   01010111 }\\
0.851 &
{\scriptsize
00000101   10110100   11111111   00010111   11110001   00111101   11111001   01010111 }\\

\hline
Coe2  &
{\scriptsize
00010100   01010001   00110000   01011100   00000000   01010000   11001110   01011111 }\\
0.860  &
{\scriptsize
00010111   00010001   11111111   01011111   00001111   01010011   11001111   01011111 }\\
\hline
\end{tabular}
\end{center}
\end{table}

\subsubsection{Spatial Distribution}
\label{gl-distOpt}

In this section, we study the spatial distribution of the six \textit{blok}.
Table \ref{tab-dist} gives the Hamming distance between these local optima. 
All the distances are lower than $64$ which
is the distance between two random solutions. 
Local optima do not seem to be randomly distributed over the landscape. Some are nearby,
 for instance GLK and Davis rules, or GLK and Coe2 rules. But Das and GLK rules, or Coe1 and Das rules
 are far away from each other.

\begin{table}
\caption{Distances between the six best local optima known}
\label{tab-dist}
\begin{center}
\begin{tabular}{lrrrrrrr}

& GLK & Davis & Das & ABK & Coe1 & Coe2 & average \\
\hline
GLK   &  0 & 20 & 62 & 56 & 39 & 34 & 28.6 \\
Davis & 20 &  0 & 58 & 56 & 45 & 42 & 33 \\
Das   & 62 & 58 &  0 & 50 & 59 & 44 & 35.4 \\
ABK   & 56 & 56 & 50 &  0 & 51 & 54 & 36.6 \\
Coe1  & 39 & 45 & 59 & 51 & 0  & 51 & 43 \\
Coe2  & 34 & 42 & 44 & 54 & 51 & 0  & 39   \\
\hline
\end{tabular}
\end{center}
\end{table}

\begin{figure}[!ht]
\begin{center}
\mbox{
  \epsfig{figure=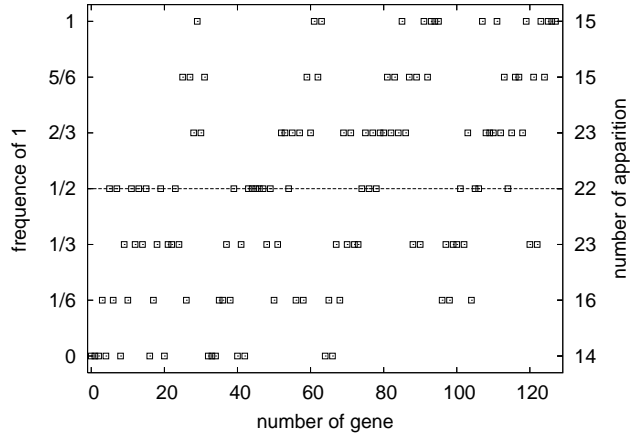,width=8.5cm,height=6cm} }
\end{center}
\caption{Centroid $C$ of the six \textit{blok}. The squares give the frequency of $1$ over the six \textit{blok} 
as function of bits position. The right column gives the number of bits of $C$ from the $128$
 which have the same frequency of $1$ indicated by the ordinate in the ordinate (left column).
\label{fig-centroide} }
\end{figure}

Figure \ref{fig-centroide} represents the centroid ($C$) of the \textit{blok}. The ordinate
is the frequency of appearance of bit value $1$ at each bit. On the right column we give the number
of bits which have the same given frequency. For six random solutions in the fitness landscape, on average
the centroid is the string with $0.5$ on the $128$ bits and the number of bits with the same frequency
 of $1$ follows the binomial law $2, 12, 30, 40, 30, 12, 2$. On the other hand, for the six best local optima,
 a large number of bits have the same value ($29$ instead of $4$ in the random case) and a smaller number of bits
 ($22$ instead of $40$ in the random case) are ``undecided'' with a frequency of $0.5$.

The local optima from the \textit{blok} are not randomly distributed over the landscape. They are all in a particular 
hyperspace of dimension $29$ defined by the following schema $S$:
\begin{center}
{ \scriptsize
000*0*** 0******* 0***0*** *****1** 000***** 0*0***** ******** *****1*1\\
0*0***** ******** *****1** ***1*111 ******** ***1***1 *******1 ***1*111
}
\end{center}

We can thus suppose that the fixed bits are useful to obtain good solutions. 
Thus, the research of a good rule would be certainly more efficient in the subspace defined by schemata $S$.
Before examining this conjecture, we are going to look at the landscape ``from the top''.

\subsubsection{Evolvability horizon}
\label{gl-profilOpt}

\begin{figure}[!ht]
\begin{center}
\begin{tabular}{ccc}
\mbox{
  \epsfig{figure=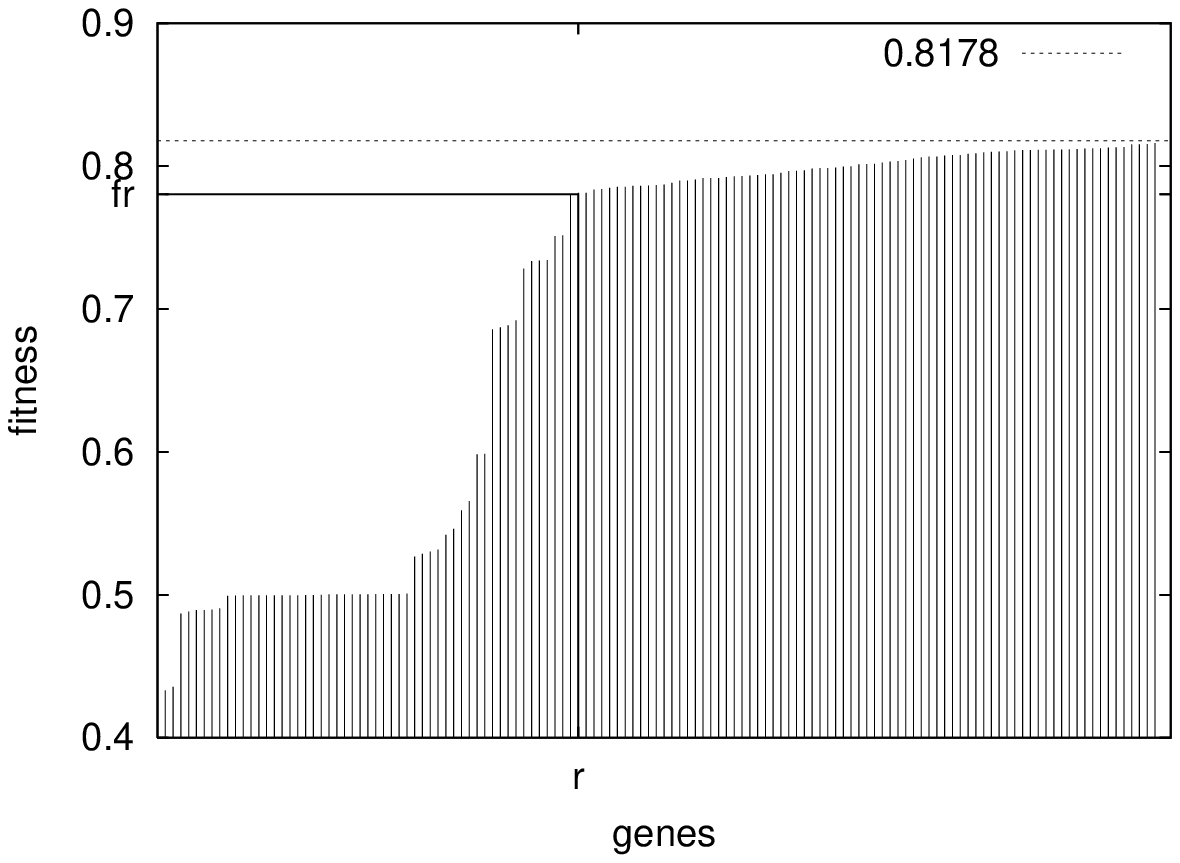,width=6.5cm,height=6cm} } & &
\mbox{
  \epsfig{figure=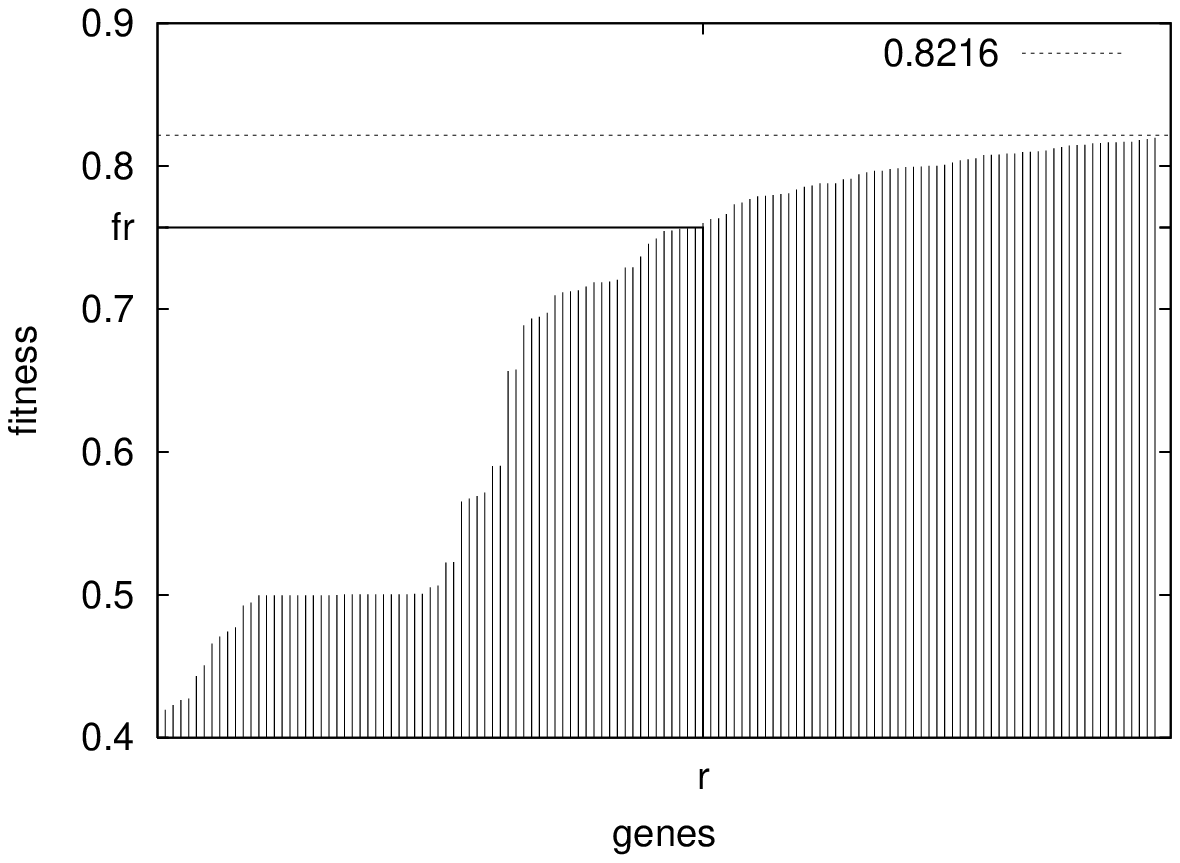,width=6.5cm,height=6cm} } \protect \\
 GLK: $r=53$, $m=0.000476$  & & Das: $r=69$, $m=0.00106$ \\
\mbox{
  \epsfig{figure=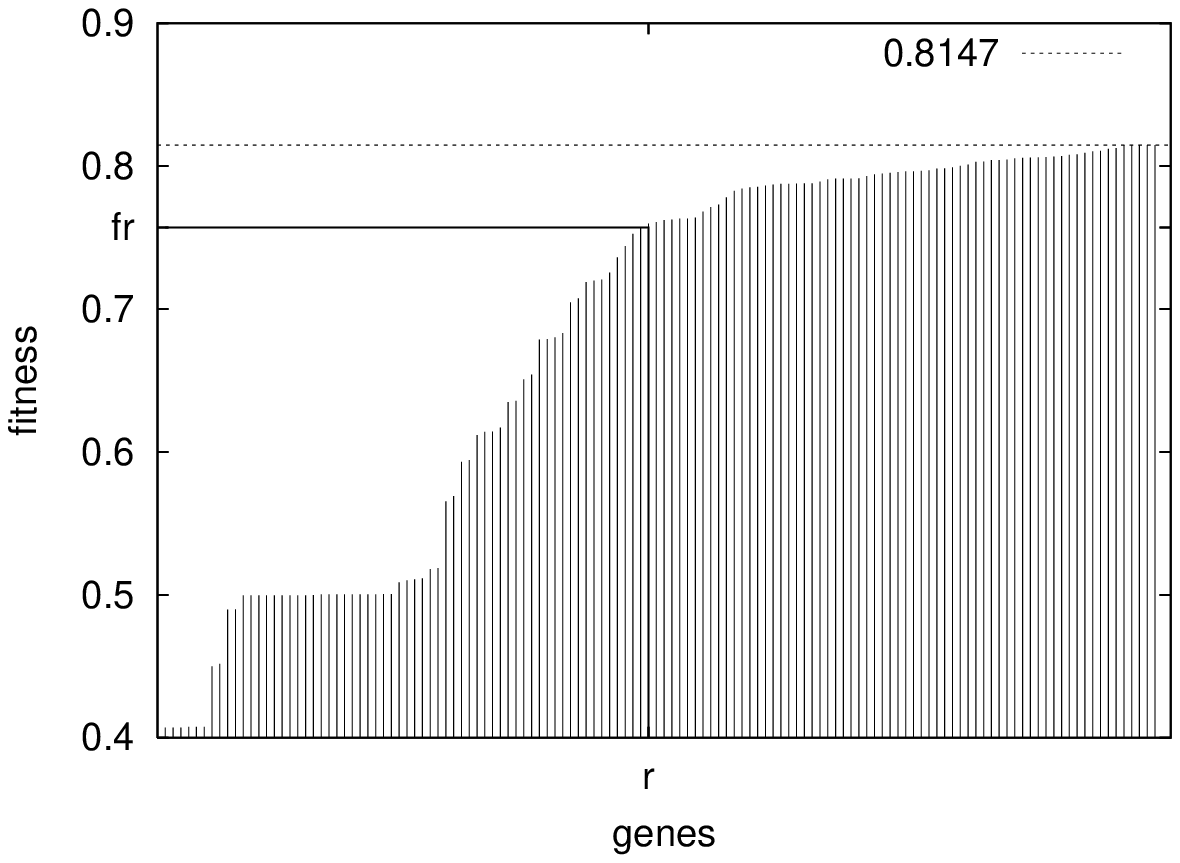,width=6.5cm,height=6cm} } & &
\mbox{
  \epsfig{figure=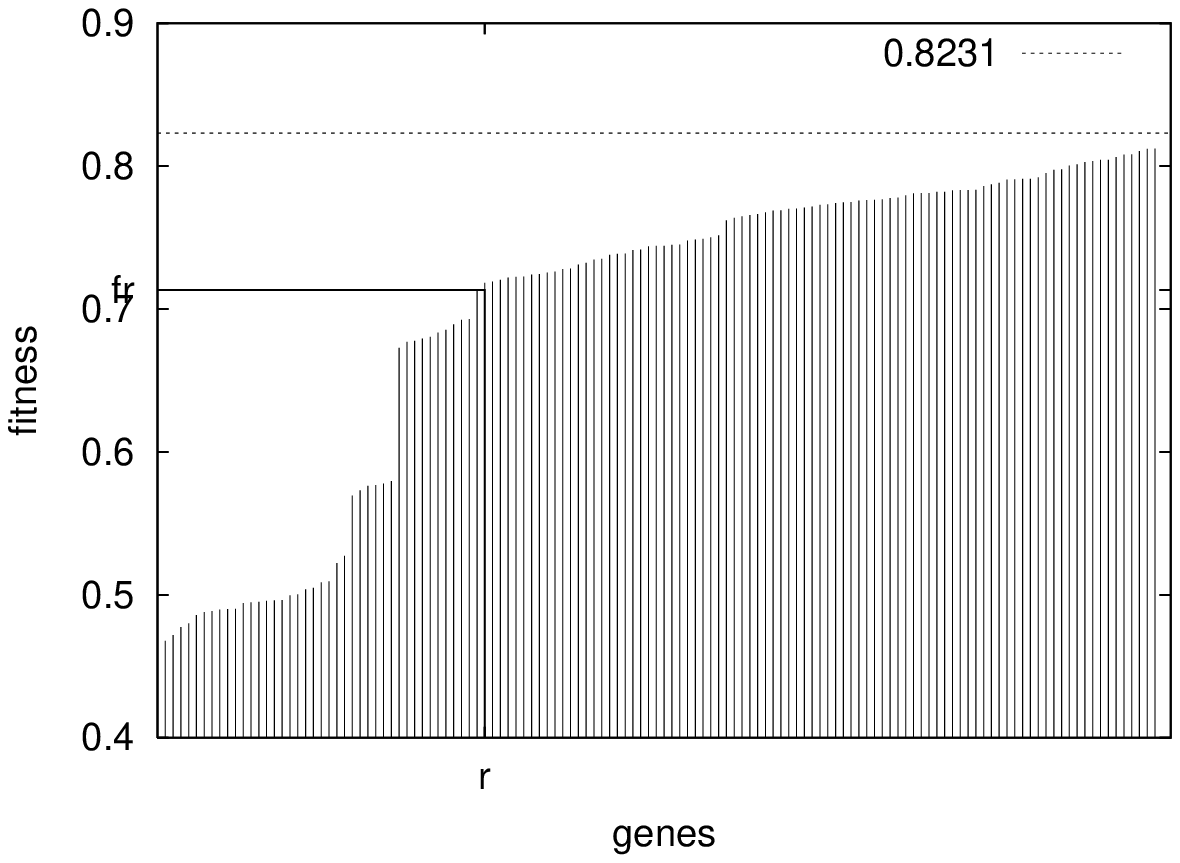,width=6.5cm,height=6cm} } \protect \\
 Davis: $r=62$, $m=0.000871$ & & ABK: $r=41$, $m=0.00114$ \\
\mbox{
  \epsfig{figure=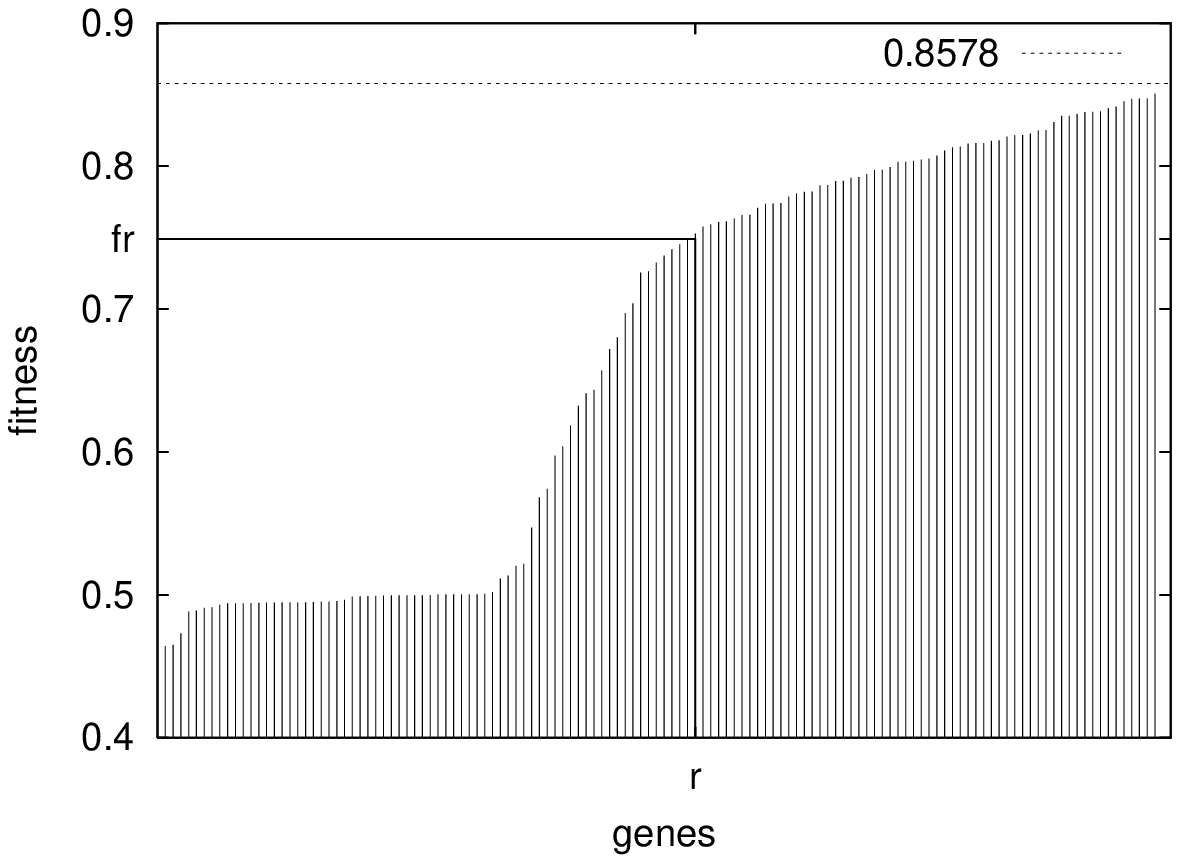,width=6.5cm,height=6cm} } & &
\mbox{
  \epsfig{figure=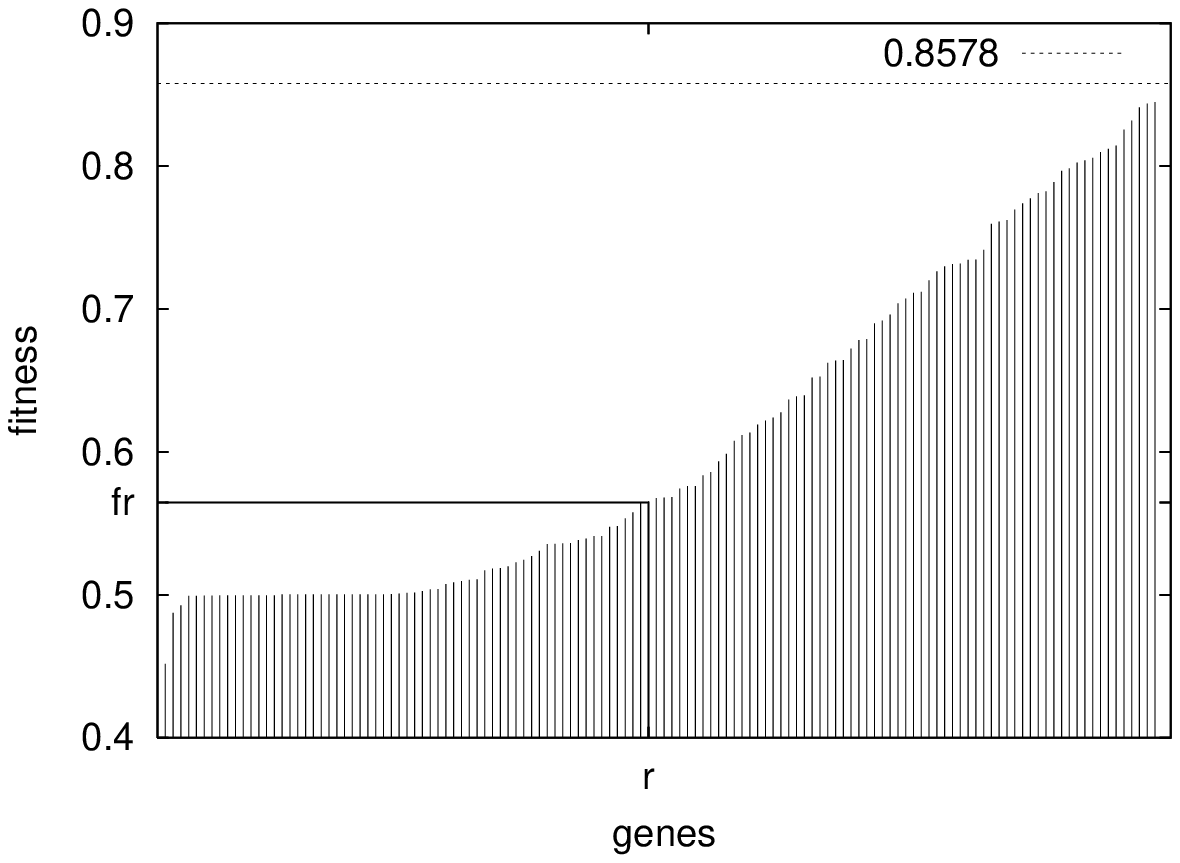,width=6.5cm,height=6cm} }  \protect  \\
Coe1: $r=68$, $m=0.00170$ & & Coe2: $r=62$, $m=0.00424$ \\
\end{tabular}
\end{center}

\caption{Evolvability horizon for the $6$ best known rules computed on binomial sample of ICs of 
size $10^4$. For each rule, the dotted horizontal line is its fitness. Row $r$ and slope $m$ (see text) 
are indicated above each figure.
\label{fig-profilOpt} }
\end{figure}

Altenberg defined evolvability as the ability to produce fitter variants \cite{kinnear:altenberg}. The 
idea is to analyze the variation in fitness between one solution and its neighbors. 
Evolvability is said positive if neighbor solutions are fitter than the solution and negative otherwise.
In this section, we define the {\it evolvability horizon} (EH) as the sequence of solutions, ordered by fitness values, which can be reached
with one bitflip from the given solution.
We obtain 
a graph with fitness values in ordinates and the corresponding neighbors in abscissa sorted by fitnesses
(see figure \ref{fig-profilOpt}).

Figure \ref{fig-profilOpt} shows the evolvability horizon of the \textit{blok}. There is no neighbor with a better fitness value than the initial rule; so, all the best known rules are local optima.
The fitness landscape has two important neutral networks at fitness $0$ ($NN_0$) and fitness $0.5$ ($NN_{0.5}$)
(see section \ref{gl-neut}). No local optimum is nearby $NN_0$; but a large part
of neighbors of local optima (around $25\%$ on average) are in $NN_{0.5}$.
As a consequence a neutral local search on $NN_{0.5}$ can potentially find a portal toward the \textit{blok}.

For each EH, there is an abscissa $r$ from which the fitness value is roughly linear.
Let $f_r$ be this fitness value, $f_{128}$ the fitness of the less sensible bit, and $m$ the slope of the curve between abscissa $r$ and $128$.
Thus, the smaller $m$ and $r$, the better the neighbors.
On the contrary, higher slope and $r$ values mean that the neighbor fitness values decay faster.\\
For example evolvability is slightly negative from GLK, as it has a low slope and a small $r$.
At the opposite, the Coe2 rule has a high slope ; this optimum is thus isolated and evolvability is strongly negative. We can imagine the space "view from GLK" flatter than the one from Coe2.

Although all EH seem to have roughly the same shape (see fig. \ref{fig-profilOpt}), we can ask whether flipping one particular bit changes the fitness in the same way.
For instance, for all the optima, flipping the first bit from '$0$' to '$1$' causes a drop in fitness.
More generally, flipping each bit, we compute the average and standard deviation of the difference in fitnesses; results are sorted into increasing average differences (see figure \ref{fig-evolOl}-a).
The bits which are the more deleterious are the ones with the smaller standard deviation, and as often as not, are in the schemata $S$. 
So, the common bits in the \textit{blok} seem to be important to find good solution: for a metaheuristic, it would be of benefit to search in the subspace defined by the schema $S$.

\clearpage

\section{Olympus Landscape}
\label{olympus}

We have seen that there are many similarity inside the \textit{blok}.
In this section we will use this feature to define the \textit{Olympus Landscape} and, to show and exploit, the relevant properties of this subspace.

\subsection{Definition}

The Olympus Landscape is a subspace of the Majority problem landscape. 
It takes its name from the \textit{Mount Olympus} which is traditionally regarded as the heavenly home of
the ancient Greek gods. Before defining this subspace we study the two natural symmetries of the majority problem.

The states 0 and 1 play the same role in the computational task ; so flipping bits in the entry of a rule and in the result have no effect on performance. In the same way, CAs can compute the majority problem according to right or left direction without changing performance. 
We denote $S_{01}$ and $S_{rl}$ respectively the corresponding operator of $0/1$ symmetry and $right/left$ symmetry. Let $x=(x_0,\ldots,x_{\lambda-1}) \in \lbrace 0,1 \rbrace^{\lambda}$ be a solution with $\lambda=2^{2r+1}$.
The $0/1$ symmetric of $x$ is $S_{01}(x) = y$ where for all $i$, $y_i = 1 - x_{\lambda-i}$.
The $right/left$ symmetric of $x$ is $S_{rl}(x) = y$ where for all $i$, $y_i = x_{\sigma(i)}$ 
with $\sigma(\sum_{j=0}^{\lambda-1} 2^{n_j}) = \sum_{j=0}^{\lambda-1} 2^{\lambda-1 - n_j}$.
The operators are commutative: $S_{rl} S_{01} = S_{01} S_{rl}$. From the $128$ bits, $16$ are invariant by $S_{rl}$ symmetry and none by $S_{01}$. Symmetry allows to introduce diversity without losing quality ; so 
evolutionary algorithm could be improved using the operators $S_{01}$ and $S_{rl}$.

We have seen that some bit values could be useful to reach a good solution (see subsection \ref{gl-profilOpt}),
and some of those are among the $29$ joint bits of the \textit{blok} (see subsection \ref{gl-distOpt}).
Nevertheless, two optima from the \textit{blok} could be distant whereas some of theirs symmetrics are closer. Here the idea is to choose for each \textit{blok} one symmetric in order to broadly maximize the number of joint bits.
The optima GLK, Das, Davis and ABK have only $2$ symmetrics because symmetrics by $S_{01}$ and $S_{rl}$ are equal. The optima Coe1 and Coe2 have $4$ symmetrics. 
So, there are $2^4.4^2=256$ possible sets of symmetrics. Among these sets, we establish the maximum number of joint bits which is possible to obtain is $51$. This ``optimal'' set contains the six \textit{Symmetrics of Best Local Optima Known} (\textit{blok$^{'}$}) presented in table \ref{tab-OptSym}. The Olympus Landscape is defined from the \textit{blok$^{'}$} as the schemata $S^{'}$ with the $51$ fixed bits above:\\
\begin{center}
{ \scriptsize
000*0*0* 0****1** 0***00** **0**1** 000***** 0*0**1** ******** 0*0**1*1\\
0*0***** *****1** 111111** **0**111 ******** 0**1*1*1 11111**1 0*01*111}\\
\end{center}

\begin{table}
\caption{Description of the 6 symmetrics of the best local optima known (\textit{blok$^{'}$}) chosen to maximize the number of joint bits.\label{tab-OptSym}}
\begin{center}
\begin{tabular}{|l|l|}
\hline
GLK$^{'}$  &
{\scriptsize
00000000   01011111   00000000   01011111   00000000   01011111   00000000   01011111 }\\
$=GLK$ & 
{\scriptsize
00000000   01011111   11111111   01011111   00000000   01011111   11111111   01011111 }\\

\hline
Das$^{'}$  &
{\scriptsize
00000000   00101111   00000011   01011111   00000000   00011111   11001111   00011111 }\\
$=Das$ &
{\scriptsize
00000000   00101111   11111100   01011111   00000000   00011111   11111111   00011111 }\\

\hline
Davis$^{'}$  &
{\scriptsize
00000000   00001111   01110011   00001111   00000000   00011111   11111111   00001111 }\\
$=S_{01}(Davis)$ &
{\scriptsize
00000000   00001111   11111111   00001111   00000000   00011111   11111111   00011111 }\\

\hline
ABK$^{'}$  &
{\scriptsize
00000000   01010101   00000000   01010101   00000000   01010101   00000000   01010101 }\\
$=S_{01}(ABK)$ &
{\scriptsize
01011111   01010101   11111111   01011111   01011111   01010101   11111111   01011111 }\\

\hline
Coe1$^{'}$  &
{\scriptsize
00000001   00010100   00110000   11010111   00010001   00001111   00111001   01010111 }\\
$=Coe1$ &
{\scriptsize
00000101   10110100   11111111   00010111   11110001   00111101   11111001   01010111 }\\

\hline
Coe2$^{'}$  &
{\scriptsize
00010100   01010101   00000000   11001100   00001111   00010100   00000010   00011111 }\\
$=S_{rl}(Coe2)$  &
{\scriptsize
00010111   00010101   11111111   11001111   00001111   00010111   11111111   00011111 }\\
\hline
\end{tabular}
\end{center}
\end{table}

\label{ol-distOpt}

The Olympus Landscape is a subspace of dimension $77$. 
All the fixed bits from schema $S$ (see section \ref{gl-distOpt}) are fixed in the schema $S^{'}$ with the same value except for the bit number $92$.

Table \ref{tab-distSym} gives the Hamming distance between the six \textit{blok$^{'}$}. All the distances are shorter than
those between the \textit{blok} (see table \ref{tab-dist}). On average, distance between the rules is 
$29.93$ for the \textit{blok$^{'}$} and $35.93$ for the \textit{blok}. Rules in the \textit{blok$^{'}$} are closer to each other with the
first four rules being closer than the two last obtained by coevolution.

\begin{table}
\caption{Distances between the symmetrics of the Best Local Optima Known (\textit{blok$^{'}$})}
\label{tab-distSym}
\begin{center}
\begin{tabular}{lrrrrrrr}

& GLK$^{'}$ & Davis$^{'}$ & Das$^{'}$ & ABK$^{'}$ & Coe1$^{'}$ & Coe2$^{'}$ & average \\
\hline
GLK$^{'}$   & 0 & 20 & 26 & 24 & 39 & 34 & 23.8 \\
Davis$^{'}$ & 20 & 0 & 14 & 44 & 45 & 42 & 27.5 \\
Das$^{'}$   & 26 & 14 & 0 & 50 & 43 & 44 & 29.5 \\
ABK$^{'}$   & 24 & 44 & 50 & 0 & 39 & 26 & 30.5 \\
Coe1$^{'}$  & 39 & 45 & 43 & 39 & 0 & 49 & 35.8 \\
Coe2$^{'}$  & 34 & 42 & 44 & 26 & 49 & 0 & 32.5 \\
\hline
\end{tabular}
\end{center}
\end{table}

\begin{figure}[!ht]
\begin{center}
\mbox{
  \epsfig{figure=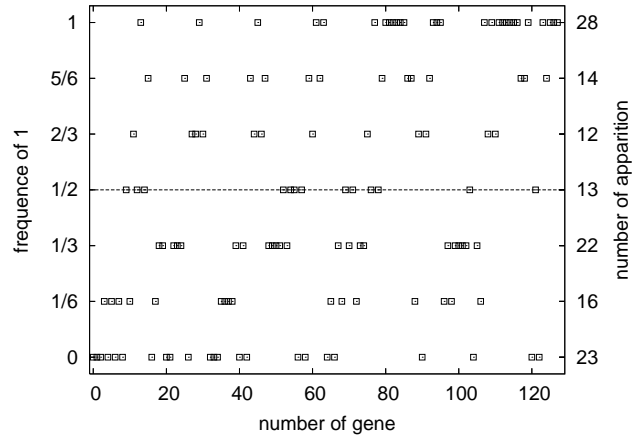,width=8.5cm,height=6cm} } 
\end{center}
\caption{Centroid of the \textit{blok$^{'}$}. The squares give the frequency of $1$ over the \textit{blok$^{'}$} as
function of bit position. The right column gives the number of bits of $C^{'}$ from the $128$  
which have the same frequency of $1$ indicated in the ordinate (left column).
\label{fig-centroide+sym} }
\end{figure}

\begin{figure}[!ht]
\begin{center}
\begin{tabular}{ccc}
\mbox{
  \epsfig{figure=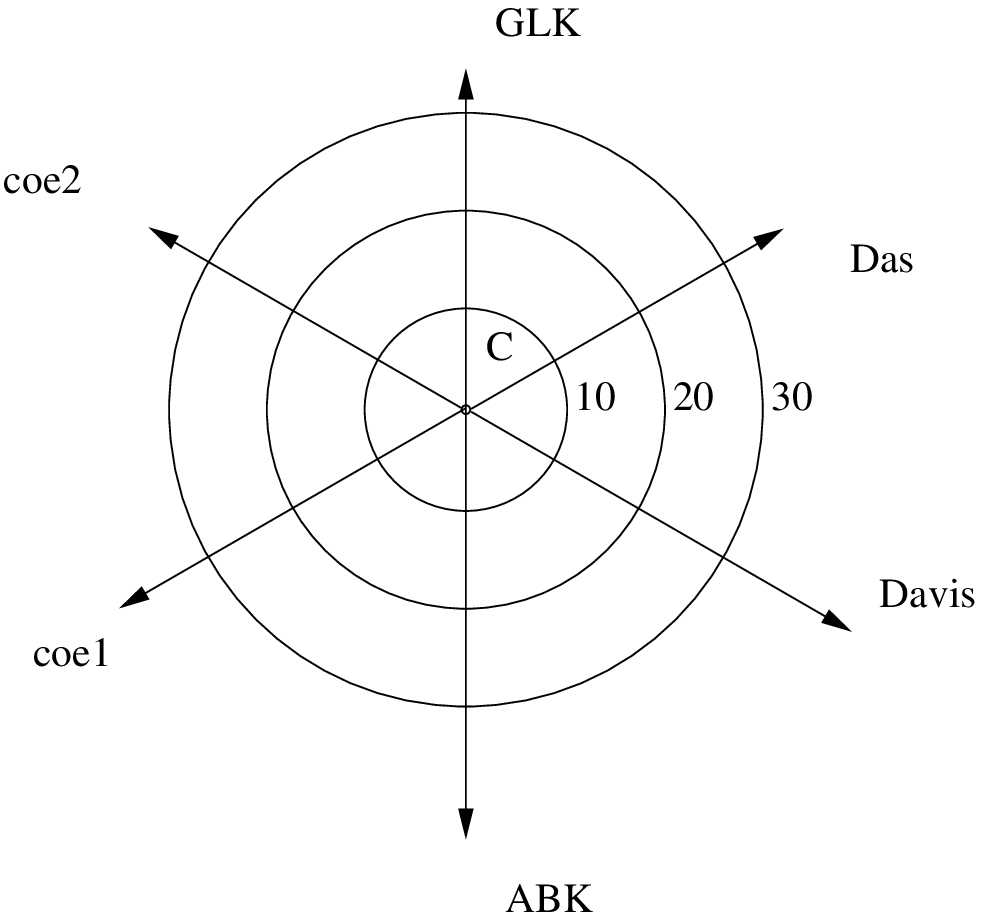,width=6cm,height=6.2cm} }  & &
\mbox{
  \epsfig{figure=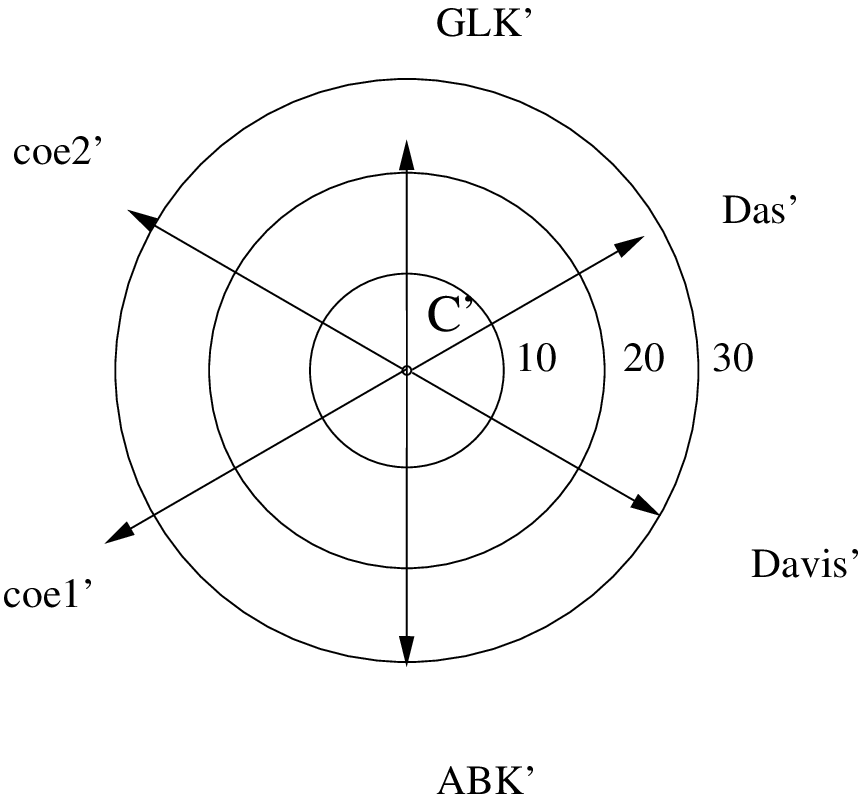,width=6cm,height=6cm} } \protect  \\
(a)   & &  (b)   \\
\end{tabular}
\end{center}
\caption{Distance between the \textit{blok} and the centroid $C$ (a) and distance between \textit{blok$^{'}$} and 
the centroid $C^{'}$ (b). \label{fig-distCent} }
\end{figure}

The centroid of the \textit{blok$^{'}$} ($C^{'}$), has
less ``undecided'' bits (13) and more fixed bits (51) than the centroid $C$ (see figure \ref{fig-centroide+sym}). Distances between $C^{'}$ and the \textit{blok$^{'}$} (see figure \ref{fig-distCent}) are shorter than the one between $C$ and the \textit{blok}. The six \textit{blok$^{'}$} are more concentrated around $C^{'}$. 
Note that, although Coe1 and Coe2 are the highest local optima, they are the farthest from $C^{'},$ although above distance $38.5$ which is the average distance between $C^{'}$ and a random point in the Olympus landscape.
This suggest one should search around the centroid while keeping one's distance from it.

\begin{figure}[!ht]
\begin{center}
\begin{tabular}{ccc}

\begin{tabular}{r}
\mbox{
  \epsfig{figure=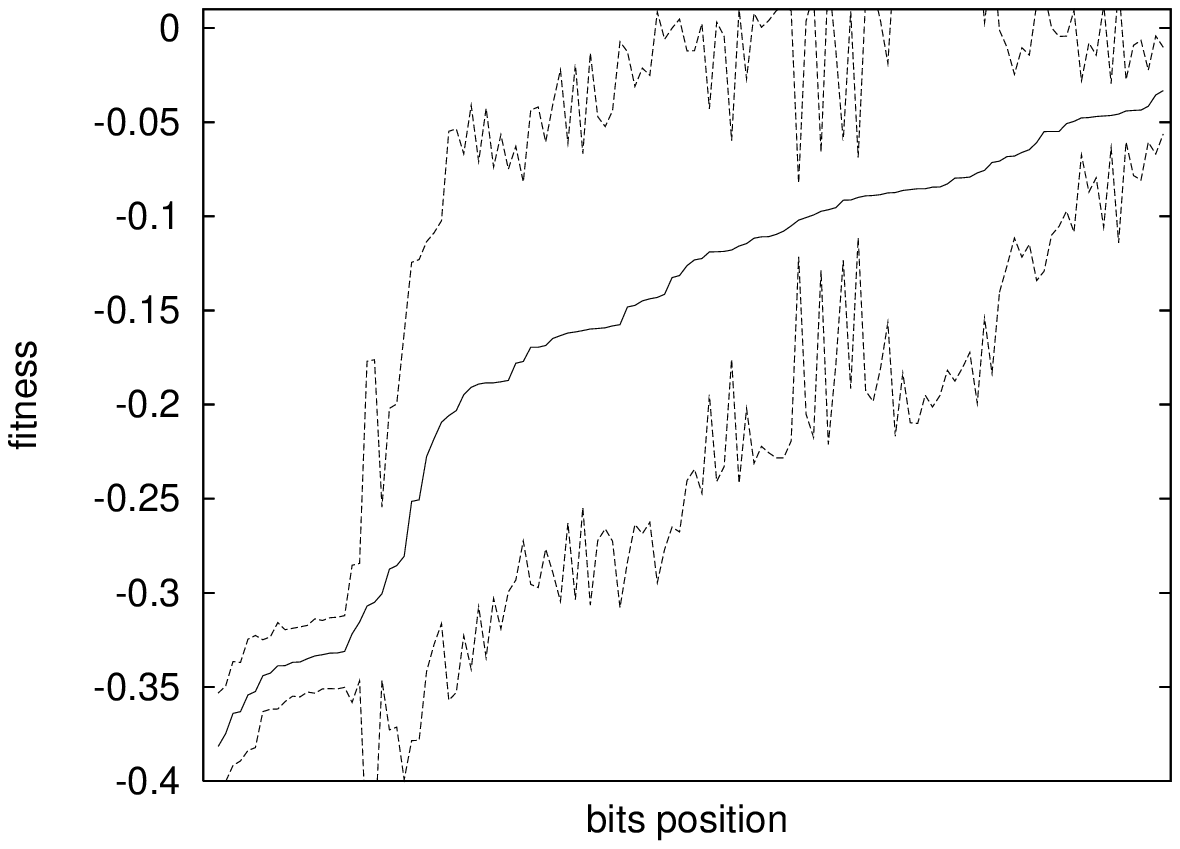,width=6.5cm,height=6cm} } \\
\mbox{
  \epsfig{figure=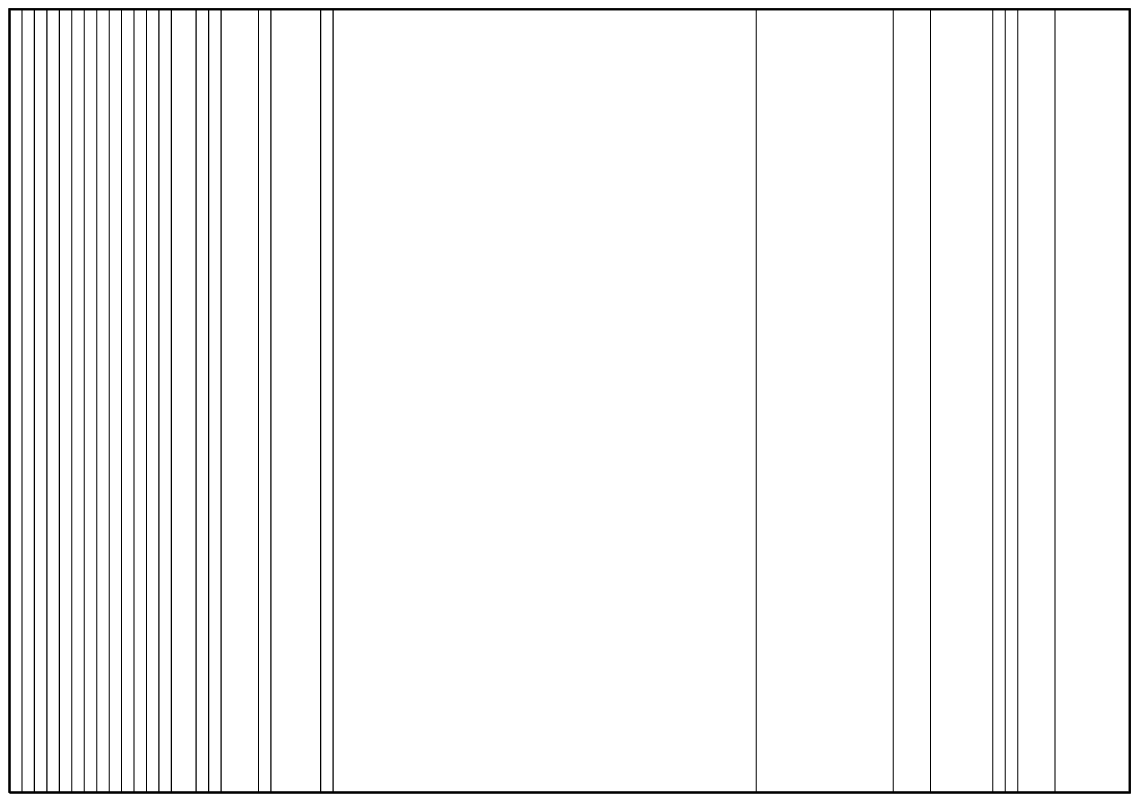,width=5.70cm,height=1cm} } \\
\end{tabular} & & 

\begin{tabular}{r}
\mbox{
  \epsfig{figure=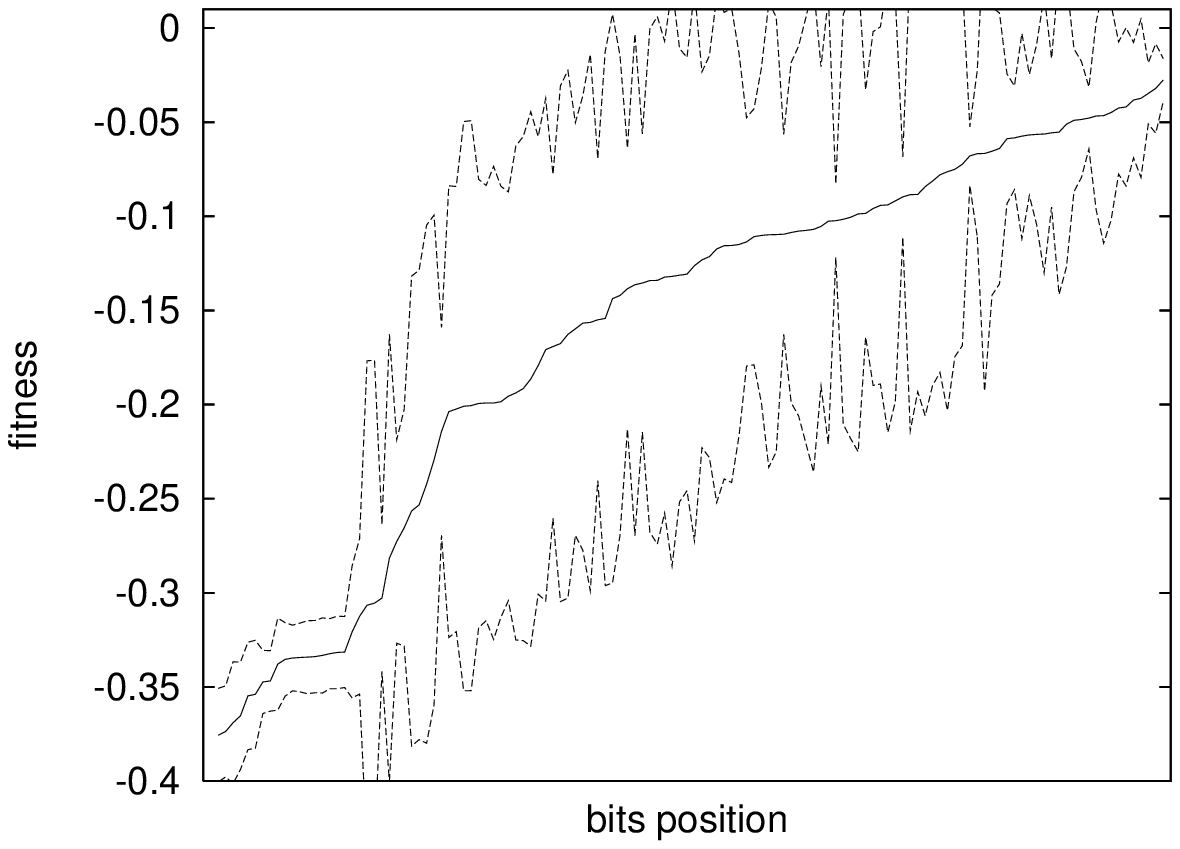,width=6.5cm,height=6cm} } \\
\mbox{
  \epsfig{figure=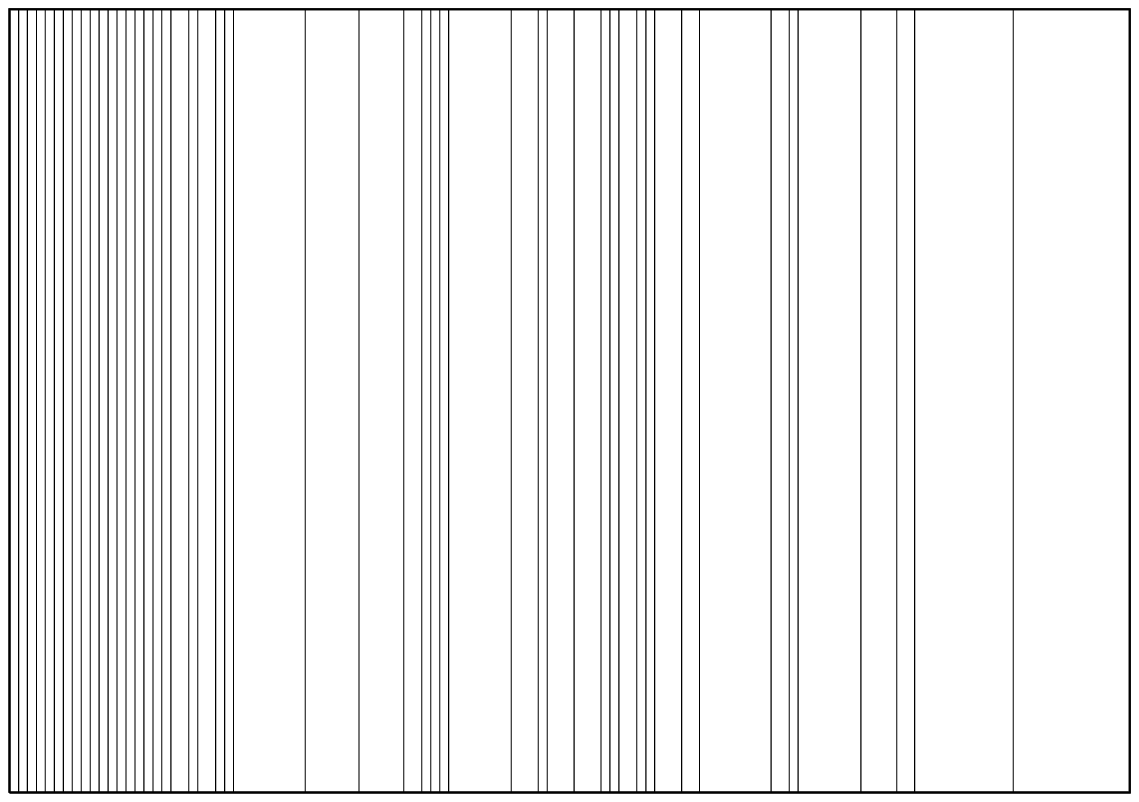,width=5.70cm,height=1cm} } \\
\end{tabular} \\

(a)   & &  (b)   \\
\end{tabular}

\end{center}
\caption{Average with standard deviation of evolvability per bit over the \textit{blok} (a) and over 
the \textit{blok$^{'}$} (b).
The boxes below the figures indicate with a vertical line which bits are in schema $S$ (a) and
schema $S^{'}$ (b).
\label{fig-evolOl} }
\end{figure}

The figure \ref{fig-evolOl}-b shows the average and standard deviation over the six \textit{blok$^{'}$} of evolvability per bit. The one over \textit{blok$^{'}$} have the same shape than the mean curve over \textit{blok}, only the standard deviation is different, on the average standard deviation is $0.08517$ for \textit{blok} and $0.08367$ for \textit{blok$^{'}$}. The Evolvability Horizon is more homogeneous for the \textit{blok$^{'}$} than for the \textit{blok}.

The Olympus contains the \textit{blok$^{'}$} which are the best rules known and 
is a subspace with unusually high fitness values easy to find as we will show in the next sections.
As such, it is a potentially interesting subspace to search. 
However, 
this does not mean that the global optimum of the whole space must
necessarily be found there.

\subsection{Statistical Measures on the Olympus Landscape}
\label{stat-oly}

In this section we present the results of the main statistical indicators restricted to the 
Olympus subspace.

\subsubsection{Density of States and Neutrality}

\begin{figure}[!ht]
\begin{center}
\begin{tabular}{ccc}
\mbox{
  \epsfig{figure=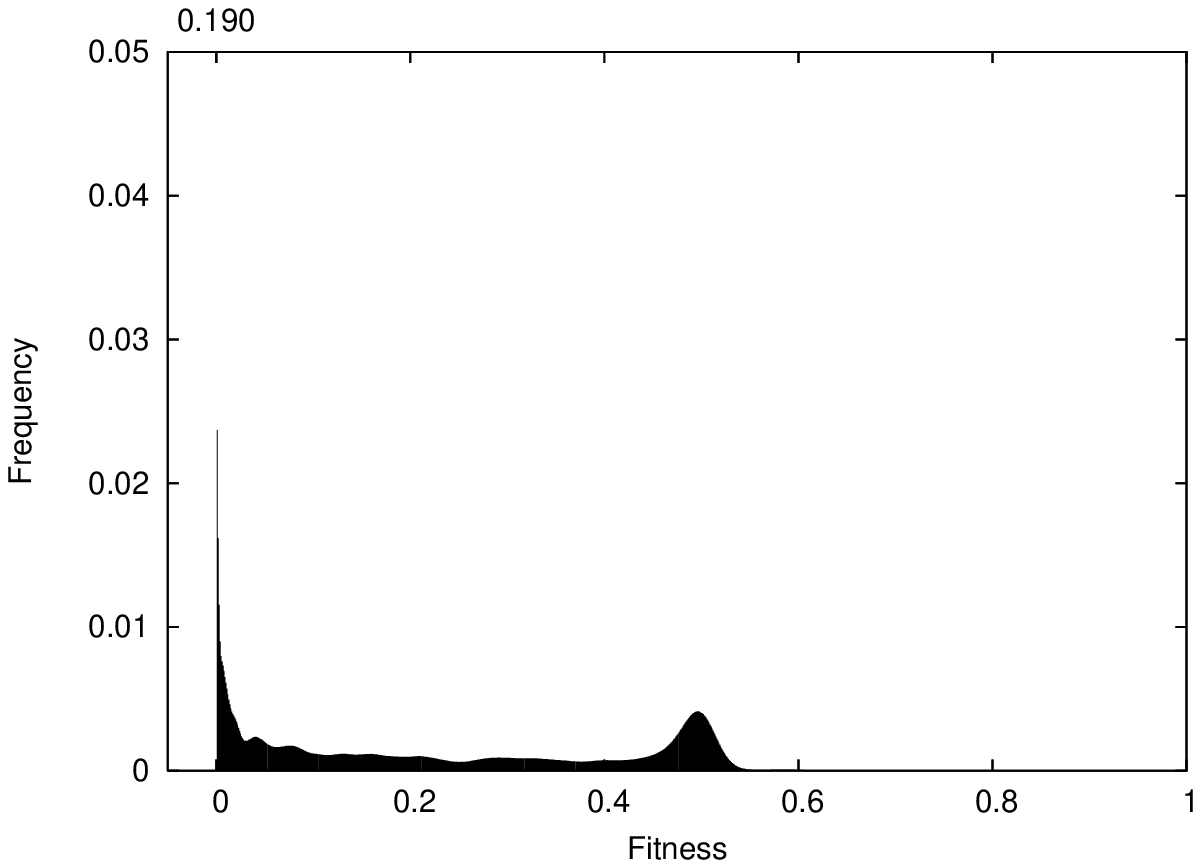,width=6.0cm,height=6cm} } & &
\mbox{
  \epsfig{figure=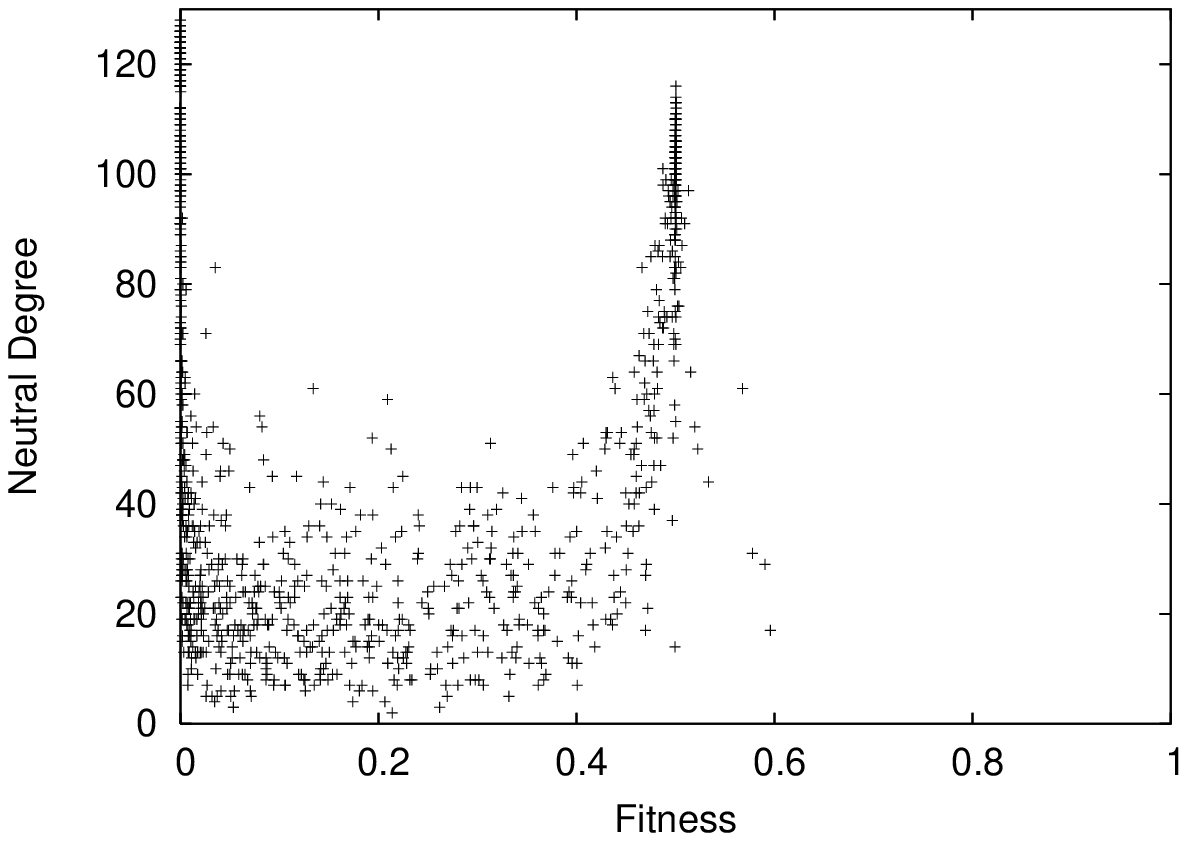,width=6.0cm,height=6cm} }  \protect  \\
(a)   & &  (b)   \\
\end{tabular}
\end{center}

\caption{Density of states (a) and Neutral degree of solutions as a function of fitness (b) on Olympus.
$10^3$ random solutions from Olympus were evaluated on a sample of ICs of size $10^4$. \label{fig-ol-dos} }
\end{figure}

Figure \ref{fig-ol-dos}-a has been obtained by sampling the space uniformly at random. The DOS is
more favorable in the Olympus with respect to the whole search space (see section \ref{dos}) although the tail of the distribution is fast-decaying beyond fitness value
$0.5$.

The neutral degree of $10^3$ solutions randomly chosen in Olympus is depicted in figure \ref{fig-ol-dos}-b.
Two important $NN$ are located around fitnesses $0$ and $0.5$ where
the neutral degree is over $80$. On average the neutral degree is $51.7$. For comparison, the average
neutral degree for NKq landscapes with $N=64$, $K=2$ and $q=2$ is $21.3$. Thus, the neutral degree 
is high on the Olympus
and this should be exploited to design metaheuristics fitting the problem.

\subsubsection{Fitness Distance Correlation}
\label{ol-fdc}

FDC has been calculated on a sample of $4000$ randomly chosen
solutions belonging to the Olympus.
Results are summarized in table \ref{tab-fdc-ol}.
The first six lines of this table reports the FDC where 
distance is calculated from one particular solution in the \textit{blok$^{'}$}. 
The line before last reports FDC where distance is computed from the nearest optimum for each individual belonging to the sample. The last line is the FDC, relative to euclidean distance, to the centroid $C^{'}$.
Two samples of solutions were generated: \textit{Osample}, where solutions are randomly chosen in the Olympus and \textit{Csample}, where each bit of a solution has probability to be '$1$' according to the centroid distribution.

\vspace*{0.2cm}

\begin{table}[!ht]
\caption{FDC where 
distance is calculated from one particular solution in the \textit{blok$^{'}$}, the nearest, or from the centroid of \textit{blok$^{'}$}. 
Two samples of solutions of size $10^4$ were generated: \textit{Osample} and \textit{Csample}.}
\label{tab-fdc-ol}
\begin{center}
\begin{tabular}{lcc}
& \textit{Osample}  & \textit{Csample} \\
\hline
GLK$^{'}$       & -0.15609 & -0.19399 \\
Davis$^{'}$     & -0.05301 & -0.15103 \\
Das$^{'}$       & -0.09202 & -0.18476 \\
ABK$^{'}$       & -0.23302 & -0.23128 \\
Coe1$^{'}$      & -0.01087 & 0.077606 \\
Coe2$^{'}$      & -0.11849 & -0.17320 \\
nearest & -0.16376 & -0.20798 \\
$C^{'}$ & -0.23446 & -0.33612 \\
\hline
\end{tabular}
\end{center}
\end{table}

\begin{figure}[!ht]
\begin{center}
\begin{tabular}{ccc}
\mbox{
  \epsfig{figure=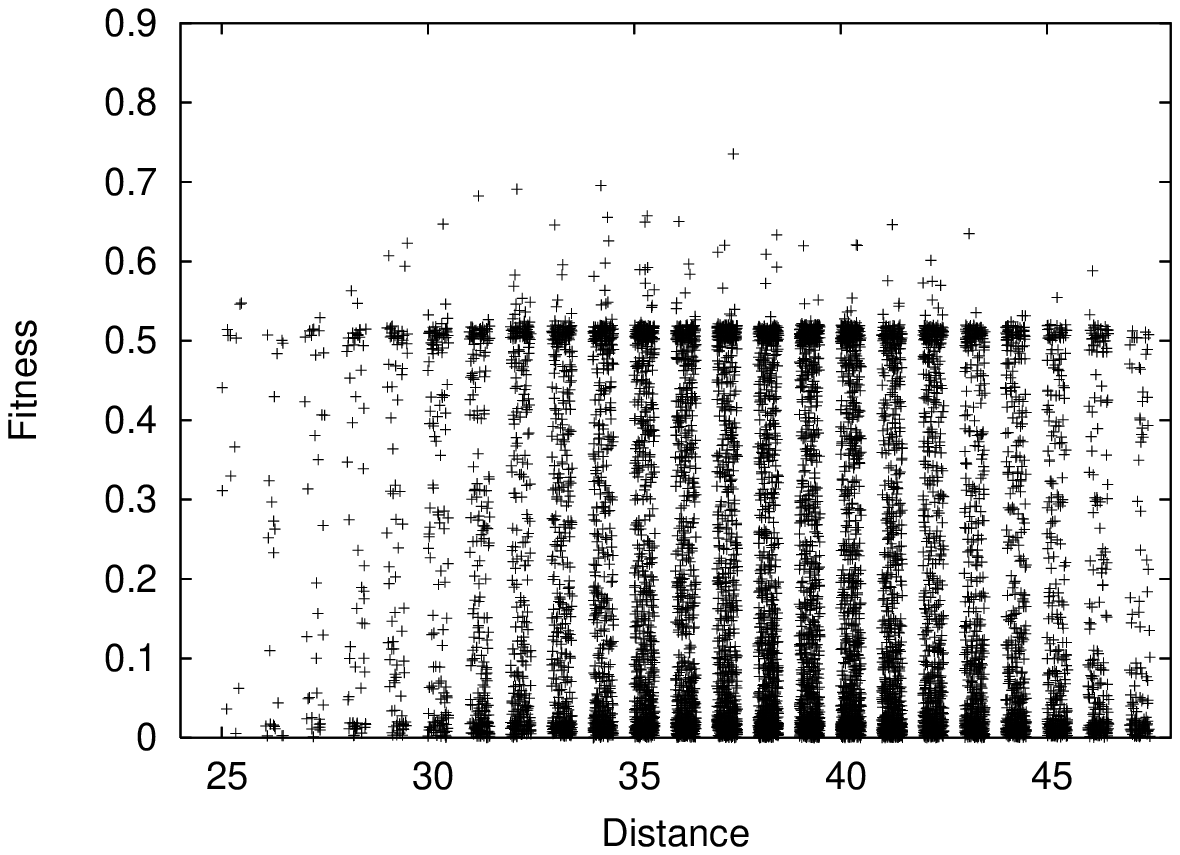,width=6.0cm,height=6cm} } & &
\mbox{
  \epsfig{figure=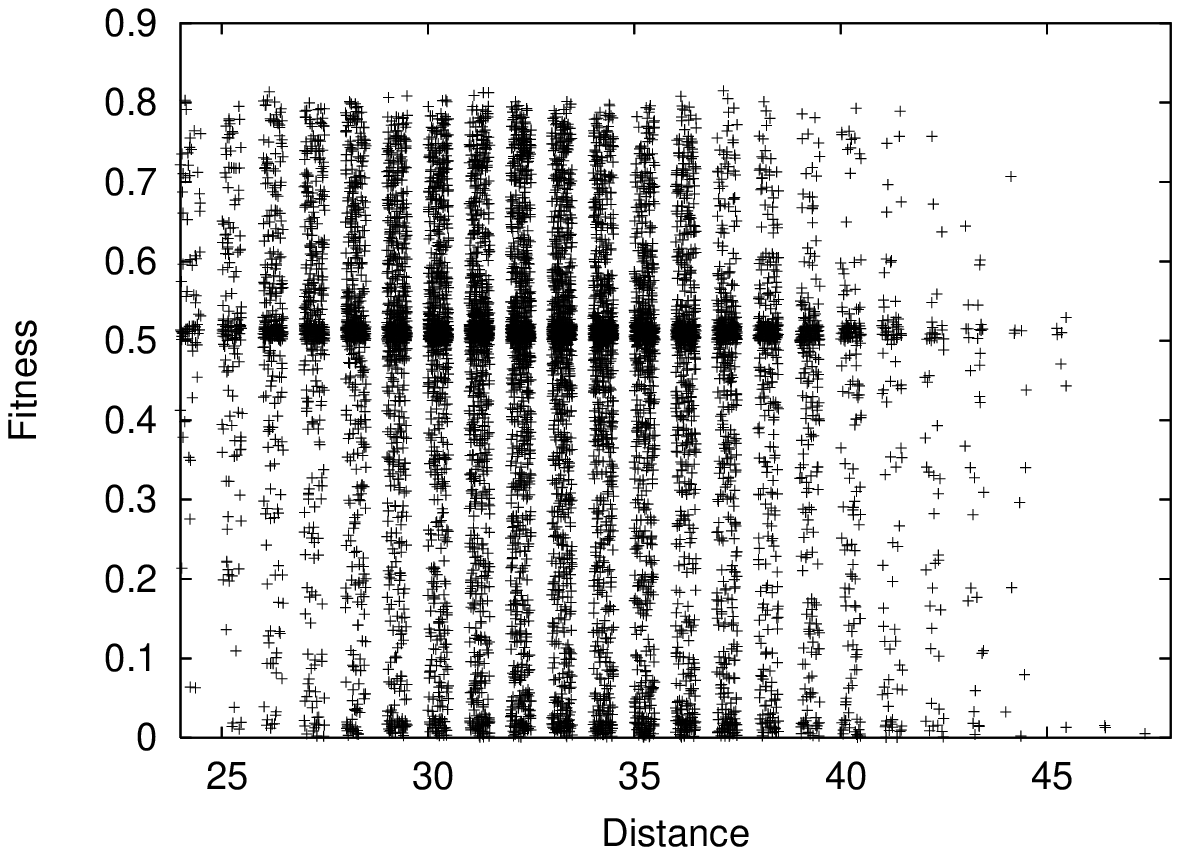,width=6.0cm,height=6cm} }  \protect  \\
(a)   & &  (b)   \\
\end{tabular}
\end{center}

\caption{FDC scatter-plot computed with euclidean distance to the centroid $C^{'}$. 
Two samples of solutions of size $10^4$ were generated: \textit{Osample} in (a) and \textit{Csample} in (b).
\label{fig-fdc} }
\end{figure}

\noindent
With the sample based on the Olympus, 
the FDC is lower, 
meaning that improvement is easier for the \textit{blok$^{'}$} than for the overall landscape (see section \ref{subsec-majLand-measures}) except for $Coe1$.
FDC with GLK$^{'}$, ABK$^{'}$, nearest, or $C^{'}$ is over the threshold $-0.15$.
For Csample, all the FDC values are lower than on Osample. Also, except for $Coe1^{'}$, the FDC is over the limit $-0.15$.
This correlation shows that fitness gives useful information to reach the local optima.
Moreover, as the FDC from the centroid is high (see also figure \ref{fig-fdc}), fitness leads to the centroid $C^{'}$. 
We can conclude that on the Olympus, fitness is a reliable guide to drive searcher toward the \textit{blok$^{'}$} and its centroid.

\subsubsection{Correlation structure analysis: ARMA model}
\label{ol-acf}
In this section we analyze the correlation structure of the Olympus landscape using the Box-Jenkins method
(see section \ref{acf}).
The starting solution of each random walk is randomly chosen on the Olympus. At each step one random bit is flipped such that the solution belongs to the Olympus
and the fitness is computed over a new sample of ICs of size $10^4$.  Random walks have length $10^4$
and the approximated two-standard-error bound used in the Box-Jenkins approach\footnote{All the statistic results have been obtained with the R programming environment (see http://r-project.org)}
is $\pm 2/\sqrt{10^4}=0.02$.

\begin{figure}[!ht]
\begin{center}
\begin{tabular}{ccc}
\mbox{
  \epsfig{figure=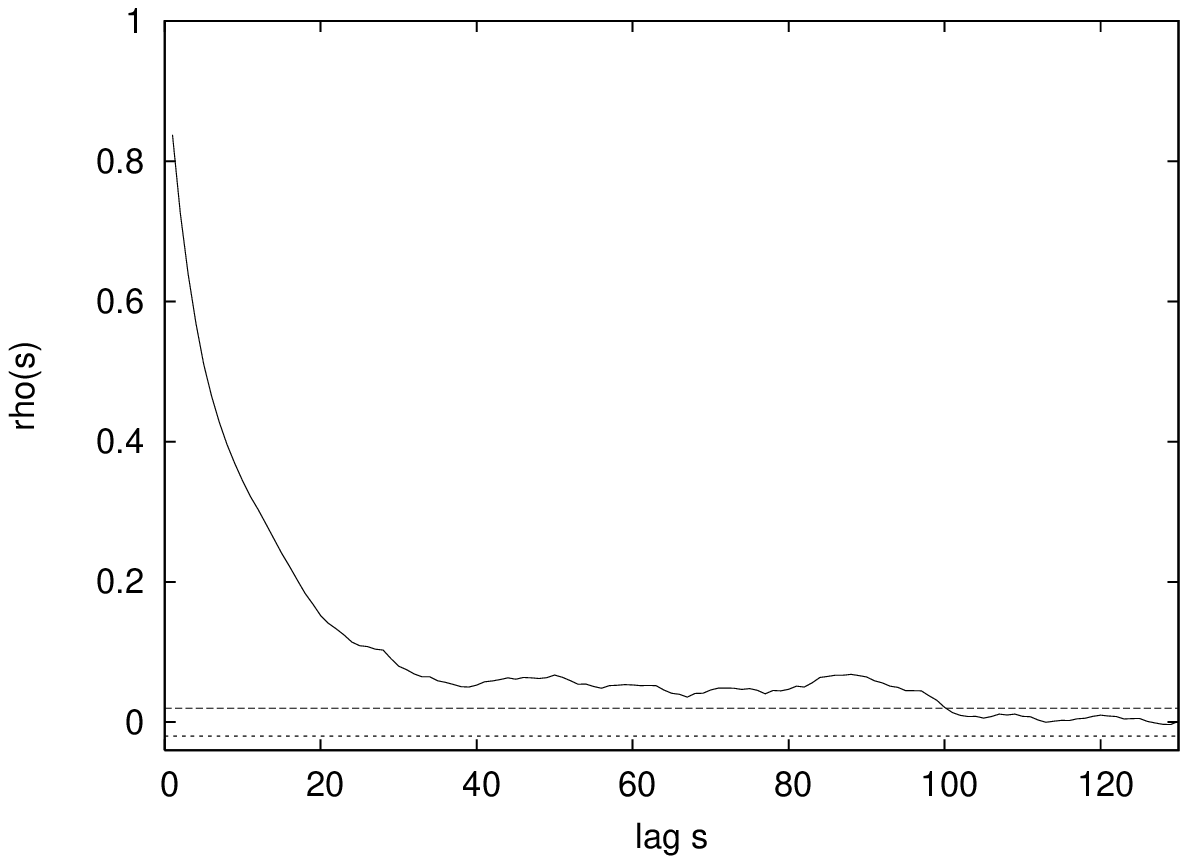,width=6.0cm,height=6cm} } & &
\mbox{
  \epsfig{figure=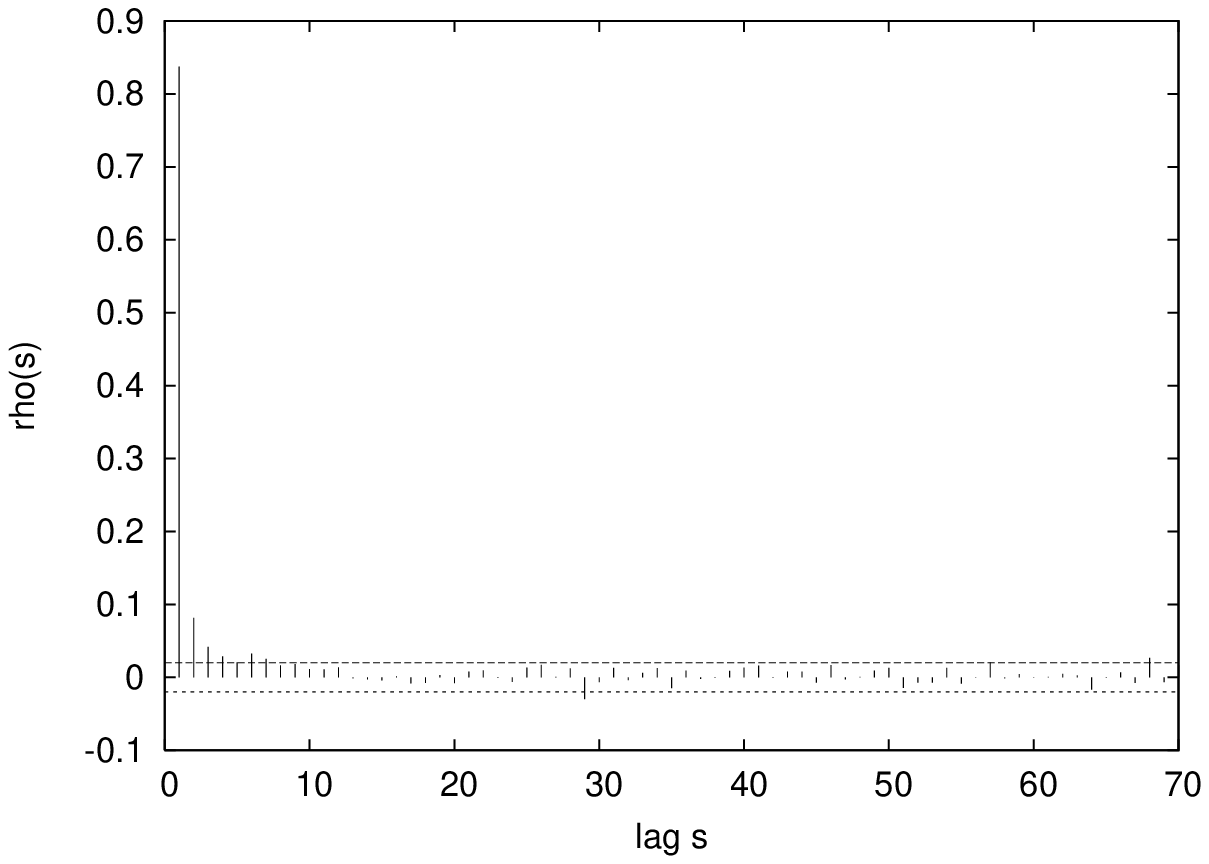,width=6.0cm,height=6cm} }  \protect  \\
(a)   & &  (b)   \\
\end{tabular}
\end{center}

\caption{Autocorrelation (a) and partial autocorrelation (b) function of random walk on Olympus.
\label{fig-autocor_unb}}
\end{figure}

\paragraph*{Identification}

Figure \ref{fig-autocor_unb} shows the estimated autocorrelation (acf) in (a) and partial autocorrelation (pacf) in (b).
The acf decreases quickly. The first-order autocorrelation is high $0.838$ and 
it is of the same order of magnitude as for NK-landscapes with $N=100$ and $K=7$ \cite{hordijk96measure}. 
The acf is closed to the two-standard-error bound from lag $40$ and cuts this bound at lag $101$ which is the correlation length.
The fourth-order partial autocorrelation is close to the two-standard-error bound. The partial autocorrelation
after lag $4$ tapers off to zero. This suggests an $AR(3)$ or $AR(4)$ model. The t-test on the estimation
coefficients of both model $AR(3)$ and $AR(4)$ are significant, but p-values of Box-Jenkins test show that 
residuals are not white noises. Thus, we tried to fit an $ARMA(3,1)$ model.
The last autoregressive coefficient $\alpha_{3}$ of the model is close to non-significant.
In order to decide the significance of this coefficient, we extracted the sequence of the $980$ first steps of the walk and estimated the model again. The t-test of $\alpha_3$ drops to $0.0738$. So $\alpha_3$ is non significant and not necessary.
We thus end up with an $ARMA(2,1)$ model.

\paragraph*{Estimation}
The results of the $ARMA(2,1)$ model estimation is:
$$
\begin{array}{lclclclclcl}
y_t & = & 0.00281 & + & 1.5384  y_{t-1} & - & 0.5665 y_{t-2} & + & \epsilon_t & - & 0.7671 \epsilon_{t-1} \\
    &   & (20.4)  &   & (32.6)          &   & (13.7)         &   &            &   & (18.1) \\
\end{array}
$$
where $y_t = f(x_t)$. The t-test statistics of the significance measure are given below the coefficients
in parentheses: they are all significant.

\begin{figure}[!ht]
\begin{center}
\begin{tabular}{ccc}
\mbox{
  \epsfig{figure=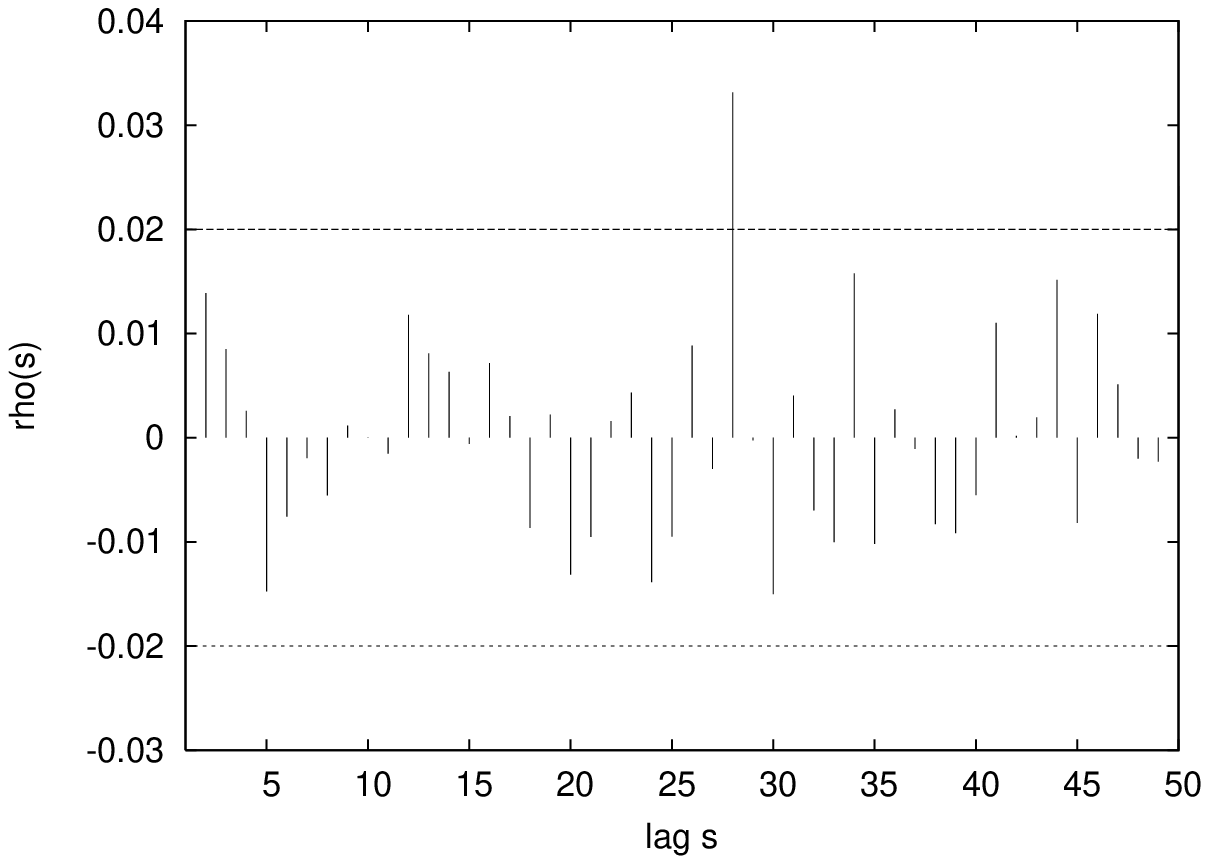,width=6.0cm,height=6cm} } & &
\mbox{
  \epsfig{figure=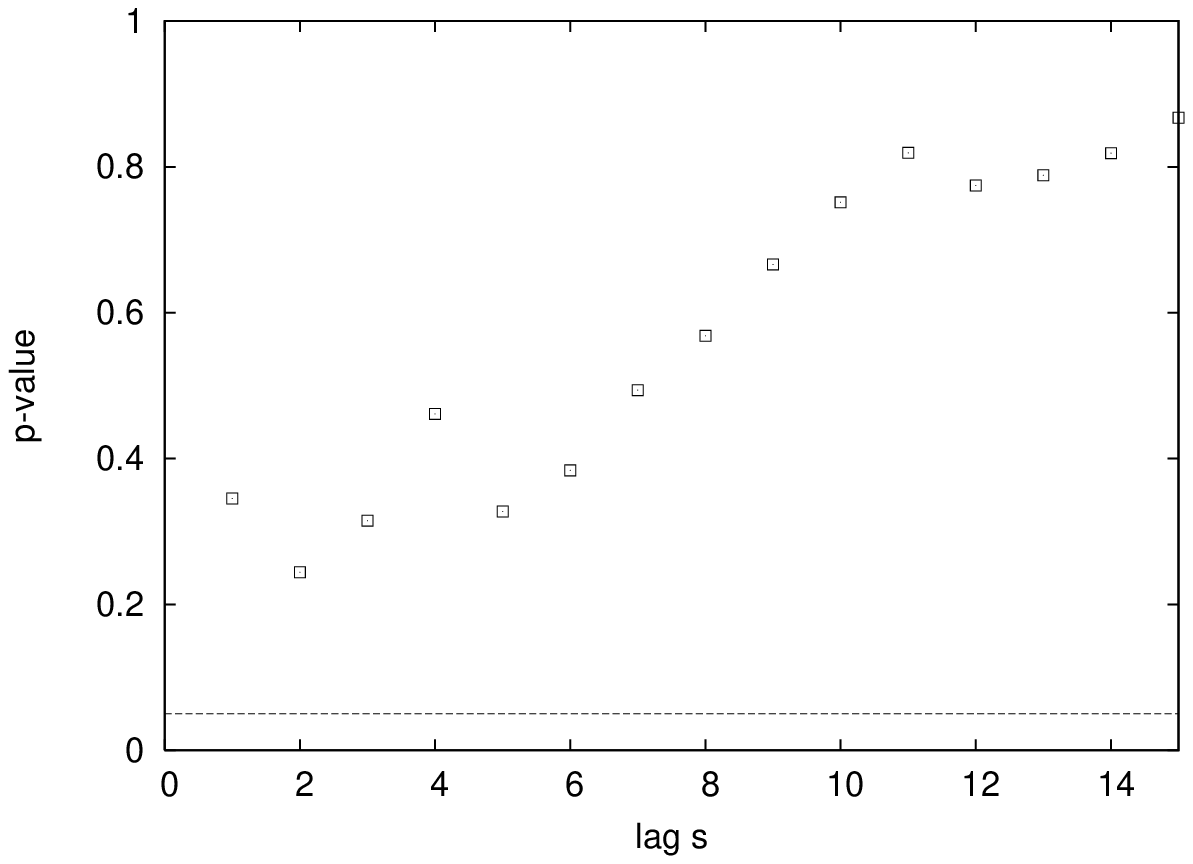,width=6.0cm,height=6cm} } \protect  \\
(a)   & &  (b)   \\
\end{tabular}
\end{center}

\caption{Autocorrelation function of residuals (a) and p-value of Ljung-Box statistic (b) for model $ARMA(2,1)$.
\label{fig-residus_unb} }
\end{figure}

\paragraph*{Diagnostic checking}
For the $ARMA(2,1)$ model estimation, the Akaide Information Criterion (aic) is $-16763.63$ and the variance 
of residuals is \linebreak $Var(\epsilon_t) = 0.01094$. Figure \ref{fig-residus_unb} shows the residuals
autocorrelation and p-value of Box-Jenkins test for the estimates of the $ARMA(2,1)$ model. The acf of residuals
are all well within the two-standard-error bound expected for $h=28$. So, the residuals are not correlated.
The p-value of Box-Jenkins test are quite good over $0.25$. The residuals can be considered as white noises.\\
The value of R-square $\bar{R}^2 = 0.7050$ is high and higher in comparison to the synchronizing-CA 
problem \cite{hord-97} where $\bar{R}^2$ is equal to $0.38$ and $0.35$. 

We can conclude that the model $ARMA(2,1)$ accurately fits the fitness values given by random walks over the Olympus Landscape. The high correlation shows that a local search heuristic is adequate to find good rules on the Olympus. An autoregressive model of size $2$ means that we need two steps to predict the fitness value; so, as suggested by Hordijk, it would be possible to construct more efficient local search taking into account this information. The moving average component has never been found in other landscape fitness analysis. What kind of useful information does it give? Maybe information on nature of neutrality. Future work should study those models in more detail.

\subsubsection{Fitness Cloud and NSC}
\label{ol-fc}

\begin{figure}[!ht]
\begin{center}
\begin{tabular}{cc}
\mbox{
  \epsfig{figure=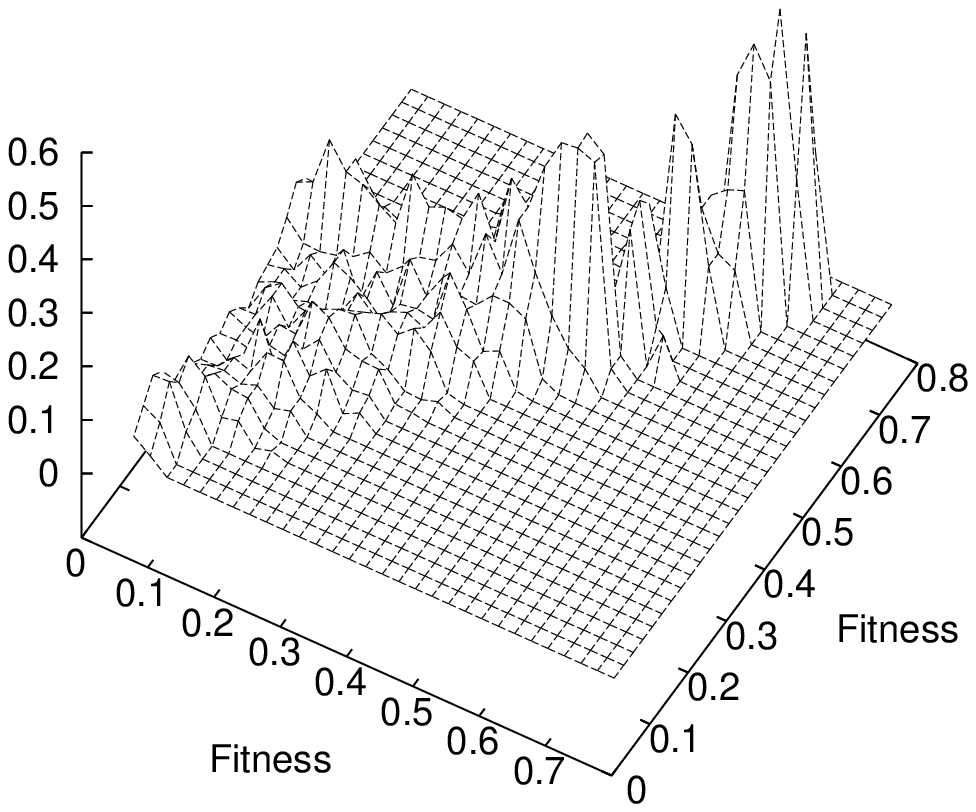,width=6cm,height=6cm} } &
\mbox{
  \epsfig{figure=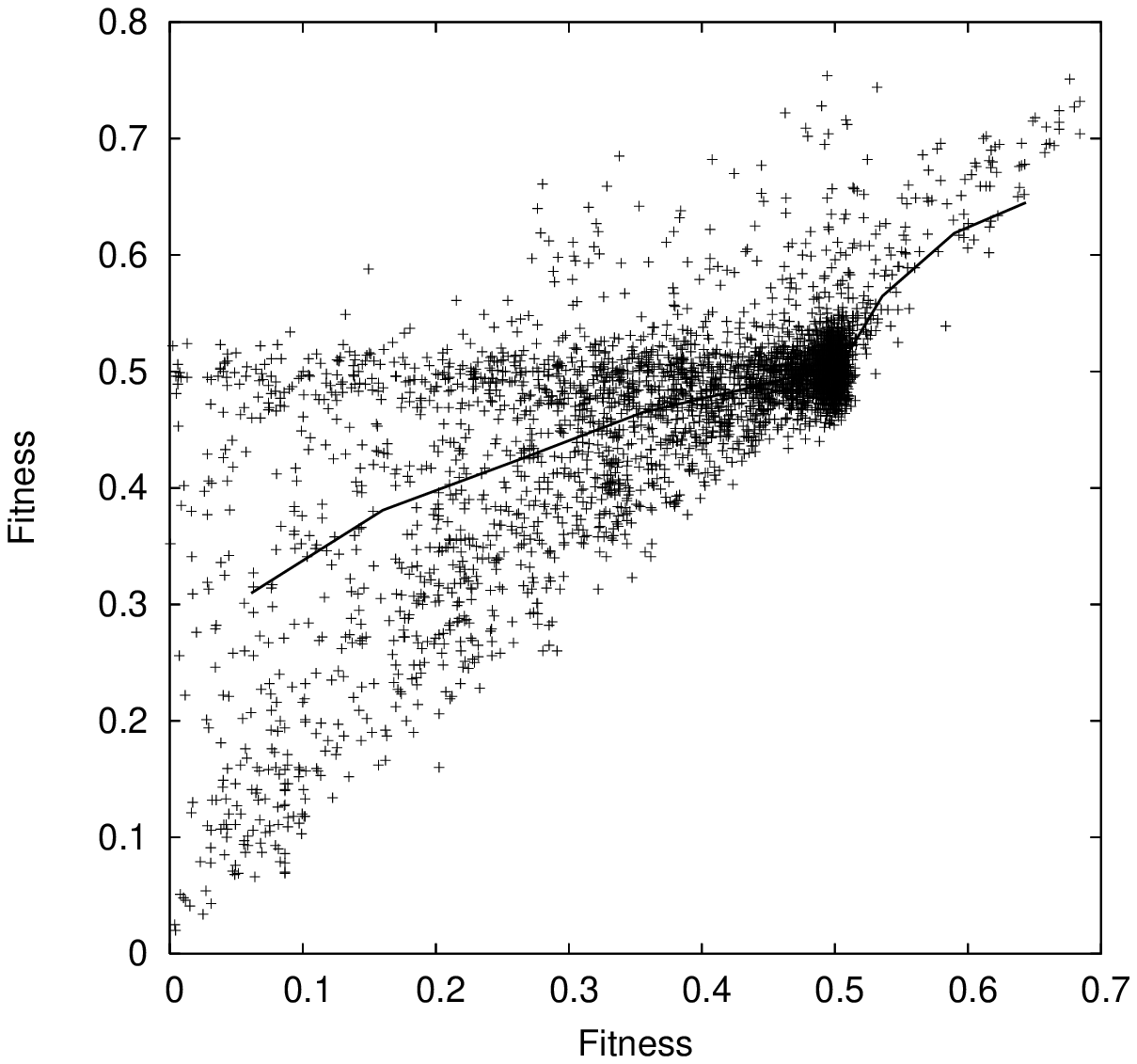,width=6cm,height=6cm} } 
\end{tabular}
\end{center}
\caption{Fitness Cloud and Scatterplot used to calculate the NSC value on Olympus.
\label{nsc-ol-scatt} }
\end{figure}

Figure \ref{nsc-ol-scatt} shows the scatterplot and the segments
$\{S_1, S_2, ..., S_{m-1}\}$ used to calculate the NSC on the Olympus (see section \ref{subsec-maj-land}).
No segment has a negative slope, 
it seems easy for a local search heuristic to reach fitness values close to $0.6$.
A comparison of this fitness cloud
with the one shown in figure \ref{nsc-whole} (where the whole
fitness landscape was considered, and not only the Olympus)
is illuminating: if the whole fitness landscape is considered,
then it is ``hard'' to find solutions with fitness up to $0.5$ ; 
on the other hand,
if only solutions belonging to the Olympus are considered,
the problem becomes much easier : 
it is now ``easy'' to access to solutions with fitness greater than $0.5$.

\subsection{Genetic Algorithms on the Olympus Landscape}
\label{ag}

In this section, we use different implementations of a genetic algorithm to
confirm our analysis of the Olympus and to find good rules to
solve the Majority problem. All implementations are based on a simple GA
used by Mitchell \textit{et al.} in \cite{mitchelletal94a}.

A population of $200$ rules is used and fitness is computed by the success
rate on unbiased sample of ICs. New individuals are first evaluated on
sample of size $10^3$. At each generation, a new sample of size $10^3$ is
generated. If an individual stays in the population during $n$ generations,
its fitness is computed from a sample of size $10^3 n$ which corresponds to
the cumulative sample of ICs. In all cases, initialization and mutation are
restricted to the Olympus. 
In order to obtain, on average, one bit mutation per individuals on Olympus, the mutation probability per bit is $1/77$. 
One point crossover is used with rate $0.6$.
We use three implementations of this GA : the \textit{Olympus based} GA
(oGA), the \textit{Centroid based} GA (cGA) and the \textit{Neutral based}
GA (nGA). The oGA allows to test the usefulness of searching in the
Olympus, the cGA tests the efficiency of searching around the centroid and
the nGA exploits the considerable neutrality of the landscape.

\paragraph*{Initial population}
For 'Olympus' and 'Neutral' based GAs, the initial population is randomly
chosen in the Olympus. For cGA, the initial population is generated according to the centroid: probability to have '$1$' at a given bit position is given by $C^{'}$ value at the same
position. In the same way, if one bit is mutated, its new value is generated
according to $C^{'}$.

\paragraph*{Selection schema}
oGA and cGA both use the same selection scheme as in  Mitchell \textit{et al.}. The top of $20$
\% of the rules in the population, so-called \textit{elite rules}, are
copied without modification to the next generation and the remaining $80$
\% for the next generation were formed by random choice in the
elite rules. The selection scheme is similar to the ($\mu + \lambda$)
selection method. The nGA uses tournament selection of size $2$. It
takes into account the neutrality of the landscape: if the fitnesses of two
solutions are not statistically different using the t-test of $95$ \%
of confidence, we consider that they are equal and choose the individual
which is the more distant from the centroid $C^{'}$; this choice allows to
spread the population over the neutral network. Otherwise the best individual is
chosen. nGA uses elitism where the top $10$ \% of different
rules in the population are copied without modification.

\begin{figure}[!ht]
\begin{center}
\begin{tabular}{ccc}
\mbox{
   \epsfig{figure=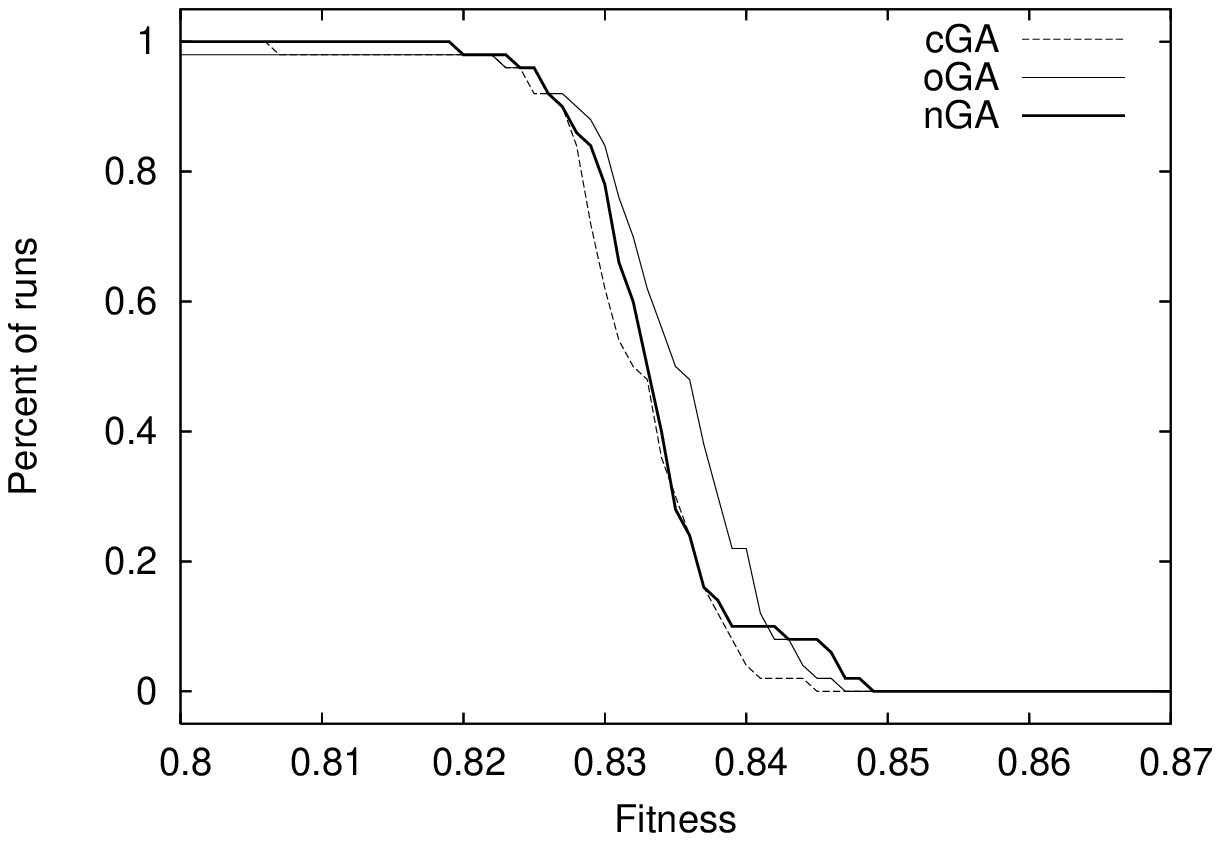,width=6.0cm,height=6cm} } & & \mbox{
   \epsfig{figure=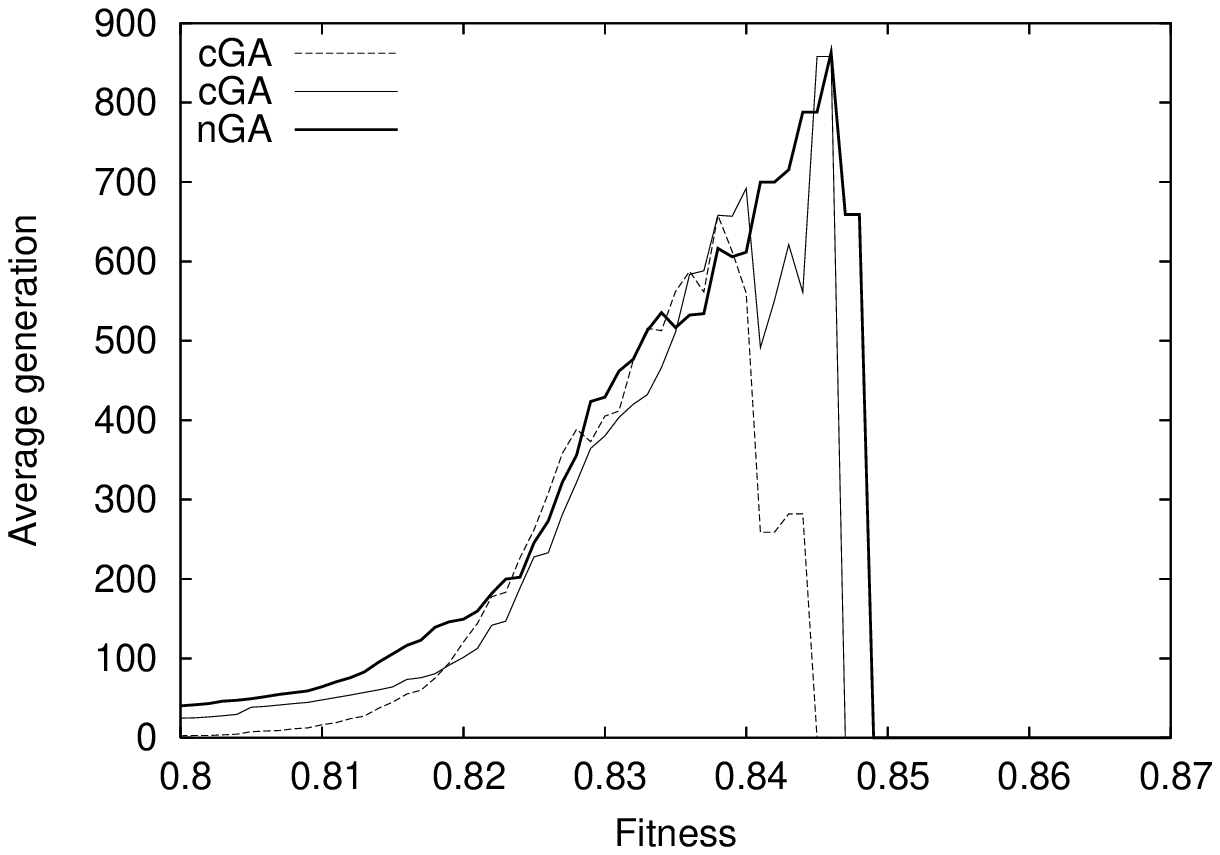,width=6.0cm,height=6cm} }  \protect  \\
(a)   & &  (b)   \\
\end{tabular}
\end{center}

\caption{Statistics percentage of runs (a) and number of generations (b) for the evolutionary emergence of CAs with 
performances exceeding various fitness thresholds.
\label{fig-threshold} } \end{figure}

\paragraph*{Performance}
Each GA run lasts $10^3$ generations and $50$ independent runs were
performed. For each run, we have performed post-processing. At each
generation, the best individuals are evaluated on new sample of $10^4$ ICs and
the average distance between all pairs of individuals is computed. Best and average
performances with standard deviation are reported in table \ref{tab-ga}. We
also computed the percentage of runs which are able to reach a given fitness
level and the average number of generations to reach this threshold (see
figure
\ref{fig-threshold}).

\begin{table}[!ht]
\caption{GA performances computed on sample of size 
$10^4$.\label{tab-ga} }

\begin{center}
\begin{tabular}{lccc}
GA & Average & Std Deviation & Best \\
\hline
oGA   & $0.8315$ & $0.01928$ & $0.8450$ \\
cGA   & $0.8309$ & $0.00575$ & $0.8432$\\
nGA   & $0.8323$ & $0.00556$ & $0.8472$\\
\hline
\end{tabular}
\end{center}
\end{table}

\begin{figure}[!ht]
\begin{center}
\mbox{
   \epsfig{figure=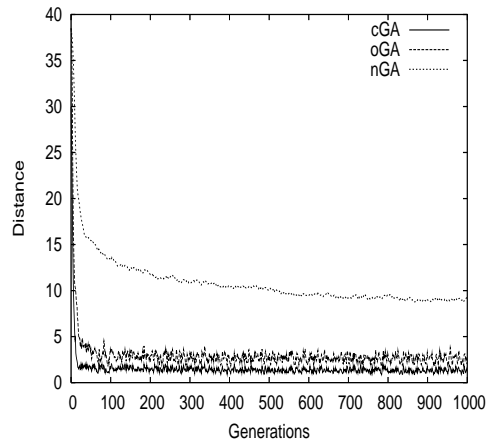,width=6.5cm,height=6cm} } \end{center}
\caption{Average hamming distance between individuals of population as a 
function of generation.
\label{fig-ga} }
\end{figure}

All GAs have \textit{on average} better performances than the 
optima find by human or by genetic programming.
As expected, searching in the Olympus is useful to find good rules. All the GAs have
nearly the same average performances. However, 
standard deviation of 'Olympus' is four times larger than standard 
deviation of 'Centroid'. As it is confirmed by the mean 
distance between individuals, the cGA quickly looses diversity (see fig. \ref{fig-ga}).
On the other hand, 'Neutral' GA keep genetic diversity during runs. 
Figure \ref{fig-threshold} shows that for the most interesting threshold 
over $0.845$, 'Neutral' have more runs able to overcome the threshold 
($3/50$) than 'Olympus' ($1/50$) or 'Centroid' ($0/50$).
Even though we cannot statistically compare the best performance of different 
GAs, the best rule was found by the nGA with performance of 
$0.8472$ to be compared to the second best rule $Coe1$.

These experimental results using GAs confirm that it is easy to find good rules in the Olympus Landscape.
During all the $50$ independent runs, we find a lot of different CAs with performance over $0.82$: $3642$ for oGA, $1854$ for cGA and $11437$ for nGA.
A 'low' computational effort is needed to obtain such CAs. 
A run takes about 8 hours on PC at 2 GHz. 
Taking the neutrality into account allows to maintain 
the diversity of the population and increases the chance to reach rules with 
high performance.

\section{Conclusions}
\label{concl}
Cellular automata are capable of universal computation and their time
evolution can be complex and unpredictable.  We have studied
CAs that perform the computational Majority task. This task is a good example of the phenomenon of emergence
in complex systems is. In this paper we have taken an interest in the
reasons that make this particular fitness landscape a difficult one. The
first goal was to study the landscape as such, and thus it is ideally
independent from the actual heuristics used to search the space. However, a
second goal was to understand the features a good search technique for this
particular problem space should possess. We have statistically quantified in
various ways some features of the landscape 
and the degree of difficulty of optimizing. 
The neutrality of the landscape is high, and the neutral network topology is not completely random.
The main observation was that 
the landscape has a considerable number of points with
fitness $0$ or $0.5$ which means that
investigations based on sampling techniques on the whole landscape are unlikely to give
good results. 

In the second part we have studied the landscape
\textit{from the top}. Although it has been proved that no CA can perform the task
perfectly, six efficient CAs for the majority task have been found either by
hand or by using heuristic methods, especially evolutionary computation. 
Exploiting similarities between these CAs and symmetries in the landscape, we have defined the
\textit{Olympus} landscape as a subspace of the Majority problem landscape
which is regarded as the "heavenly home" of the six \textit{symmetric of best local optima known} (\textit{blok$^{'}$}). 
Then, we have measured several properties of the Olympus landscape and 
we have compare with those of the full landscape,
finding that there are less solutions with fitness $0$.
FDC shows that fitness is a reliable guide to drive a searcher toward the \textit{blok$^{'}$} and its centroid.
An $ARMA(2,1)$ model has been used to describe the fitness/fitness correlation structure.
The model indicates that local search heuristics are adequate for finding good rules.
Fitness clouds and nsc confirm that it is easy to reach solutions with fitness higher than $0.5$.
Although it is easier to find relevant CAs in this subspace than in the complete landscape, 
there are structural reasons that prevents a searcher from finding overfitted GAs in the Olympus.
Finally, we have studied the
dynamics and performances of three Genetic Algorithms on the Olympus in order to
confirm our analysis and to find efficient CAs for the
Majority problem with low computational effort.

Beyond this particular optimization problem, the method presented in this
paper could be generalized. Indeed, in many optimization problems, several
efficient solutions are available, and we can make good use of this set to
design an "Olympus subspace" in the hope of finding better solutions
or finding good solutions more quickly.

\bibliographystyle{elsart-num}


\end{document}